\documentclass[10pt,twocolumn,journal]{IEEEtran}
\usepackage{xcolor}
\usepackage{float}
\usepackage{mathrsfs}
\usepackage{amsfonts}
\usepackage{amssymb}
\usepackage{amsthm}
\usepackage{epsfig}
\usepackage{graphicx}
\usepackage{bm}
\usepackage{amsmath,latexsym,amsthm,array,algorithm,algpseudocode,booktabs}
\usepackage{cite}
\usepackage{subfig}
\usepackage{multirow}
\usepackage{makecell}
\usepackage{subfig}
\usepackage{longtable}
\usepackage{threeparttable}
\ifCLASSINFOpdf
\else
\fi
\begin{document}
\title{\textbf{When Autonomous Systems Meet Accuracy and Transferability through AI: A Survey}}
\author{Chongzhen Zhang\IEEEauthorrefmark{1}, 
        Jianrui Wang\IEEEauthorrefmark{1},
        Gary G. Yen,
        Chaoqiang Zhao,\\
        Qiyu Sun,
        Yang Tang,
        Feng Qian,
        and J\"{u}rgen Kurths

\thanks{This work was supported by the National Key Research and Development Program of China under Grant 2018YFC0809302, the National Natural Science Foundation of China under Grant Nos. 61988101, 61751305, 61673176, by the Programme of Introducing Talents of Discipline to Universities (the 111 Project) under Grant B17017.}
\thanks{\IEEEauthorrefmark{1}These two authors contributed equally to this work.}
\thanks{C. Zhang, J. Wang, C. Zhao, Q. Sun, Y. Tang and F. Qian are with the Key Laboratory of Advanced Control and Optimization for Chemical Process, Ministry of Education, East China University of Science and Technology, Shanghai 200237, China (e-mail: yangtang@ecust.edu.cn (Y. Tang)).}
\thanks{G. Yen is with the School of Electrical and Computer Engineering, Oklahoma State University, Stillwater, OK 74075 USA (e-mail: gyen@okstate.edu).}
\thanks{J. Kurths is with the Potsdam Institute for Climate Impact Research, 14473 Potsdam, Germany, and with the Institute of Physics, Humboldt University of Berlin, 12489 Berlin, Germany (e-mail: juergen.kurths@pik-potsdam.de).}

% <-this % stops a space
%\thanks{Manuscript received April 19, 2005; revised January 11, 2007.}
}

\markboth{When Autonomous Systems Meet Accuracy and Transferability through AI: A Survey} {Shell
\MakeLowercase{\textit{et al.}}: Bare Demo of IEEEtran.cls for
Journals}
\maketitle

\begin{abstract}
%\boldmath
With widespread applications of artificial intelligence (AI), the capabilities of the perception, understanding, decision-making and control for autonomous systems have improved significantly in the past years. When autonomous systems consider the performance of accuracy and transferability, several AI methods, like adversarial learning, reinforcement learning (RL) and meta-learning, show their powerful performance. Here, we review the learning-based approaches in autonomous systems from the perspectives of accuracy and transferability. Accuracy means that a well-trained model shows good results during the testing phase, in which the testing set shares a same task or a data distribution with the training set. Transferability means that when a well-trained model is transferred to other testing domains, the accuracy is still good. Firstly, we introduce some basic concepts of transfer learning and then present some preliminaries of adversarial learning, RL and meta-learning. Secondly, we focus on reviewing the accuracy or transferability or both of them to show the advantages of adversarial learning, like generative adversarial networks (GANs), in typical computer vision tasks in autonomous systems, including image style transfer, image super-resolution, image deblurring/dehazing/rain removal, semantic segmentation, depth estimation, pedestrian detection and person re-identification (re-ID). Then, we further review the performance of RL and meta-learning from the aspects of accuracy or transferability or both of them in autonomous systems, involving pedestrian tracking, robot navigation and robotic manipulation. Finally, we discuss several challenges and future topics for using adversarial learning, RL and meta-learning in autonomous systems.
\end{abstract}


% \fi
\IEEEpeerreviewmaketitle

\section{Introduction}

Artificial intelligence (AI) has been widely used in art, government, healthcare, games and economics, due to its powerful learning ability. Especially after the representative AI algorithm AlphaGo defeated the world champion in Go games \cite{silver2017mastering}, people have been paying more attention to AI. Understanding the behavior of AI agents is very important to promote its technology \cite{rahwan2019machine}. With the rise of deep learning (DL) algorithms, the upgrading of hardwares and the availability of big data, AI technology has been making huge progress these years \cite{lecun2015deep}. Autonomous systems powered by AI, including unmanned vehicles, robotic manipulators and drones, etc, have been widely used in various industries and daily lives, such as intelligent transportation \cite{ferdowsi2019deep}, intelligent logistics \cite{mcfarlane2016intelligent} and service robots \cite{forlizzi2006service}, etc. Due to the limitations of current computer perception and decision-making technologies in terms of accuracy and transferability, autonomous systems still have much room to be improved for complex and intelligent tasks via technological development. Due to the ability of DL to capture high-dimensional data features \cite{lecun2015deep}, DL-based algorithms are widely used in the perception and decision-making tasks of autonomous systems. There are a number of typical perception and decision-making related tasks for autonomous systems, such as image super-resolution (SR) \cite{dong2014learning}, \cite{dong2015image}, image deblurring/dehazing/rain removal \cite{sun2015learning}, \cite{cai2016dehazenet}, \cite{eigen2013restoring}, semantic segmentation \cite{long2015fully}, \cite{badrinarayanan2017segnet}, depth estimation \cite{eigen2014depth}, \cite{eigen2015predicting}, pedestrian detection \cite{tian2015pedestrian}, person re-identification (re-ID) \cite{ye2017dynamic}, pedestrian tracking \cite{supancic2017tracking}, robot navigation \cite{kober2013reinforcement}, \cite{polydoros2017survey}, and robotic manipulation \cite{gupta2017learning}, \cite{faust2018prm}, etc. However, most DL-based models have good accuracy and poor transferability, i.e., they are usually effective in the testing dataset with the same data distribution or task. When a well-trained model is transferred to other datasets or real-world tasks, the accuracy usually declines drastically, which means that the transferability is poor, and thus the transferability has to be taken into account for practical applications \cite{tan2018survey}. This issue results in the fact that the current vision perception and decision-making methods cannot be used directly in actual autonomous systems. Transfer learning improves the transferability of models between different domains, i.e., a well-trained model can achieve a good accuracy when applied to other testing domains.

Recently, since adversarial learning, like generative adversarial networks (GANs), has shown its promising results in image generation, a number of GANs-based methods have been proposed and achieved breakthroughs in the above computer vision tasks \cite{zhu2017unpaired}, \cite{atapour2018real},  \cite{ledig2017photo}, \cite{kupyn2018deblurgan}, etc. In the field of AI, GANs have become more and more important due to their powerful generation and domain adaptation capabilities \cite{gui2020review}. GANs have attracted increasing attention, since they were proposed by Goodfellow \textit{et al.} \cite{goodfellow2014generative} in 2014. GAN is a generative model that introduces adversarial learning between the generator and the discriminator, in which the generator creates data to deceive the discriminator, while the discriminator distinguishes whether its input comes from real data or generated ones. The generator and discriminator are iteratively optimized in the game, and finally reach the Nash equilibrium \cite{goodfellow2016nips}. In particular, when considering a well-trained model for different datasets or real scenes, GANs can be used for domain transfer tasks by virtue of their ability to capture high-frequency features to generate sharp images \cite{csurka2017domain}. Although some learning-based models mainly focus on the aspect of accuracy \cite{dong2014learning}, \cite{long2015fully}, \cite{eigen2014depth}, GANs have demonstrated satisfactory results for various complex image fields in autonomous systems and other related fields, such as text-to-image generation \cite{reed2016generative}, \cite{xu2018attngan}, image style transfer \cite{zhu2017unpaired}, \cite{isola2017image}, SR \cite{ledig2017photo}, image deblurring \cite{kupyn2018deblurgan}, image rain removal \cite{qian2018attentive}, \cite{zhang2019image}, object detection \cite{li2017perceptual}, \cite{ehsani2018segan}, semantic segmentation \cite{zhu2017unpaired}, \cite{hong2018conditional}, \cite{sankaranarayanan2018learning}, pedestrian detection \cite{kim2019unpaired}, person re-ID \cite{deng2018image}, and video generation \cite{vondrick2016generating}, etc. 

Meanwhile, as a powerful tool for decision-making and control, reinforcement learning (RL) has been extensively studied in recent years, because it is suitable for decision-making tasks in complex environments \cite{jaradat2011reinforcement}, \cite{kohl2004policy}. However, when the input data are high-dimensional such as images, sounds and videos, it is difficult to solve the problem only with RL. With the help of deep neural networks (DNNs), deep RL (DRL), which combines the high-dimensional perceptual ability of DL with the decision-making ability of RL, has achieved promising results recently in various fields of application, such as obstacle avoidance \cite{xie2017towards}, \cite{chen2018deep}, robot navigation \cite{kahn2018self}, \cite{anderson2018vision}, robotic manipulation \cite{levine2016end}, \cite{mnih2016asynchronous}, video target tracking \cite{supancic2017tracking}, \cite{luo2019end}, games playing \cite{parisotto2015actor}, \cite{rusu2015policy}, and drug testing \cite{olivecrona2017molecular}, \cite{popova2018deep}, etc. However, DRL tends to require a large number of trials and needs to specify a reward function to define a certain task \cite{botvinick2019reinforcement}. The former is time-consuming and the latter is significantly difficult when training from scratch. In order to tackle these problems, the idea of “learn to learn”, called meta-learning emerged \cite{vilalta2002perspective}. Compared with DRL, meta-learning makes the learning methods more transferable and efficient by utilizing previous experience to guide the learning of new tasks across domains. Therefore, meta-learning methods perform well especially in environments lacking data, such as image recognition \cite{koch2015siamese}, classification \cite{finn2017model}, robot navigation \cite{zhu2017target} and robotic arm control \cite{duan2017one}, etc. 

%In the review, we will put the emphasis on a certain meta-learning method called MAML, analyze and list some specific applications of the methods \cite{yu2018one} \cite{al2017continuous}.

\begin{table}
	\centering
	\caption{Summary of abbreviations in this review}
	\begin{tabular}{cc}
		\toprule
		Abbreviation & Full Name \\
		\midrule
		AC & actor-critic\\
		AI & artificial intelligence\\
		cGANs & conditional generative adversarial networks\\
		CNNs & convolutional neural networks\\
		CycleGAN & cycle-consistent adversarial networks\\
		DL & deep learning\\
		DNNs & deep neural networks\\
		DQN & deep Q network\\
		DRL & deep reinforcement learning\\
		GANs & generative adversarial networks\\
		GAIL & generative adversarial imitation learning\\
		HR & high-resolution\\
		IRL & inverse reinforcement learning\\
		LR & low-resolution\\
		LSTM & long short-term memory\\
		MAML & model-agnostic meta-learning\\
		re-ID & re-identification\\
		RL & reinforcement learning\\
		SR & super-resolution\\
		TCN & temporal convolution network\\
		\bottomrule
	\end{tabular}
	\label{table1}
\end{table}

With the development of DL, learning-based perception and decision-making algorithms for autonomous systems have become a hot research topic. There are some reviews on autonomous systems, Tang \textit{et al.} \cite{tang2020perception} introduced the applications of learning-based methods in perception and decision-making for autonomous systems. Gui \textit{et al.} \cite{gui2020review} gave a detailed overview of various GANs methods from the perspectives of algorithms, theories and applications. Arulkumaran \cite{arulkumaran2017brief} detailed the core algorithms of DRL and the advantages of RL for visual understanding tasks. Unlike previous surveys, we focus on reviewing learning-based approaches in the perception and decision-making tasks of autonomous systems from the perspectives of accuracy or transferability or both of them. 

The organization of this review is arranged as follows. Section \ref{2} introduces transfer learning and one of its related machine learning techniques, domain adaptation. Then the basic concepts of adversarial learning, RL and meta-learning are presented. In Section \ref{3}, we survey some recent developments by exploring various learning-based approaches in autonomous systems, taking into account accuracy or transferability or both of them. In Section \ref{4}, we summarize some trends and challenges for autonomous systems. Conclusions are given in Section \ref{5}. We summarize the abbreviations in this review in Table \ref{table1}.

\begin{table*}
	\centering
    \small
	\caption{Definitions and differences between three transfer learning settings \copyright (2009) IEEE. Reprinted, with permission, from \cite{pan2009survey}}
	\begin{tabular}{ccccc}
	\toprule
		Transfer learning settings	& Source and target domains	& Source and target tasks	& Source domain labels	& Target domain labels \\
	\midrule
		inductive transfer learning	& the same/different but related	& different but related	& available/unavailable	& available\\
		transductive transfer learning	& different but related	& the same	& available	& unavailable\\
		unsupervised transfer learning	& the same/different but related	& different but related	& unavailable	& unavailable\\
    \bottomrule
	\end{tabular}
	\label{table2}
\end{table*}

\section{Preliminaries} \label{2}

Learning-based methods are used in various perception and decision-making tasks of autonomous systems, such as image style transfer, image super-resolution (SR), image deblurring/dehazing/rain removal, semantic segmentation, depth estimation, pedestrian detection, person re-ID, pedestrian tracking, robot navigation and robotic manipulation, etc. However, most traditional learning-based methods usually achieve good accuracy on the testing set with the same distribution or the same task. In recent years, with the research on the transferability of models, several typical learning-based methods are widely used, like adversarial learning and meta-learning, etc.

\subsection{Overview of the section}
Focusing on the transferability of models, transfer learning is proposed and first introduced in this section, which aims to make a well-trained model have a good transferability, i.e., the well-trained model can be transferred to other testing sets and still have a good accuracy. Then, we introduce several typical learning-based methods concentrating on improving the accuracy or transferability or both of them, including adversarial learning, RL, and meta-learning. In the perception tasks of autonomous systems, adversarial learning, like GANs, has capabilities of good accuracy or transferability or both of them. In the decision-making tasks of autonomous systems, RL and meta-learning are often used to improve the accuracy or transferability of the system.
%Moreover, meta-learning performs well than RL especially when the training data is lacking suffers from poor transferability, and therefore meta-learning is used in combination with RL to improve the transferability of the model.

\subsection{Transfer learning}
\textbf{Transfer learning.}
Transfer learning is a research topic aiming to investigate the improvement of learners from one target domain trained with more easily obtained data from source domains \cite{weiss2016survey}. In other words, the domains, tasks and distributions used in training and testing could be different. Therefore, transfer learning saves a great deal of time and cost in labeling data when encountering various scenarios of machine learning applications. According to different situations between domains, source tasks and target tasks, transfer learning can be categorized into three subsettings: inductive transfer learning, transductive transfer learning and unsupervised transfer learning \cite{pan2009survey}. The definitions and differences between these transfer learning settings can be seen in detail in Table \ref{table2}.

%Specifically, In the inductive transfer learning setting, the target task is different from the source task, whether the source and target domains are the same or not. We mention task-transfer \cite{}, feature-representation-transfer \cite{parisotto2015actor} and parameter-transfer \cite{zhu2017target} approaches in the review.
%In the transductive transfer learning setting, the source and target tasks are the same, while the source and target domains are different. We talk about task-transfer \cite{}, feature-representation-transfer \cite{} approaches in the review.
%Similar to inductive transfer learning, in the unsupervised transfer learning setting, the target task is different from but related to the source task. Moreover, there are no labeled
%data available in both source and target domains in training. In the review, we mainly discuss feature-representation-transfer \cite{}, \cite{} approaches.
%In this review, we will put emphasis on transductive transfer learning problem, where the training and testing tasks are drawn from the different but related distributions, that is, domain adaptation \cite{yang2018transductive}.

\begin{figure*}[htbp]
	\centering
	\subfloat[GAN;]
	{
		\begin{minipage}[t]{0.48\textwidth}
			\centering
			\includegraphics[width=0.9\textwidth, height = 3.8cm]{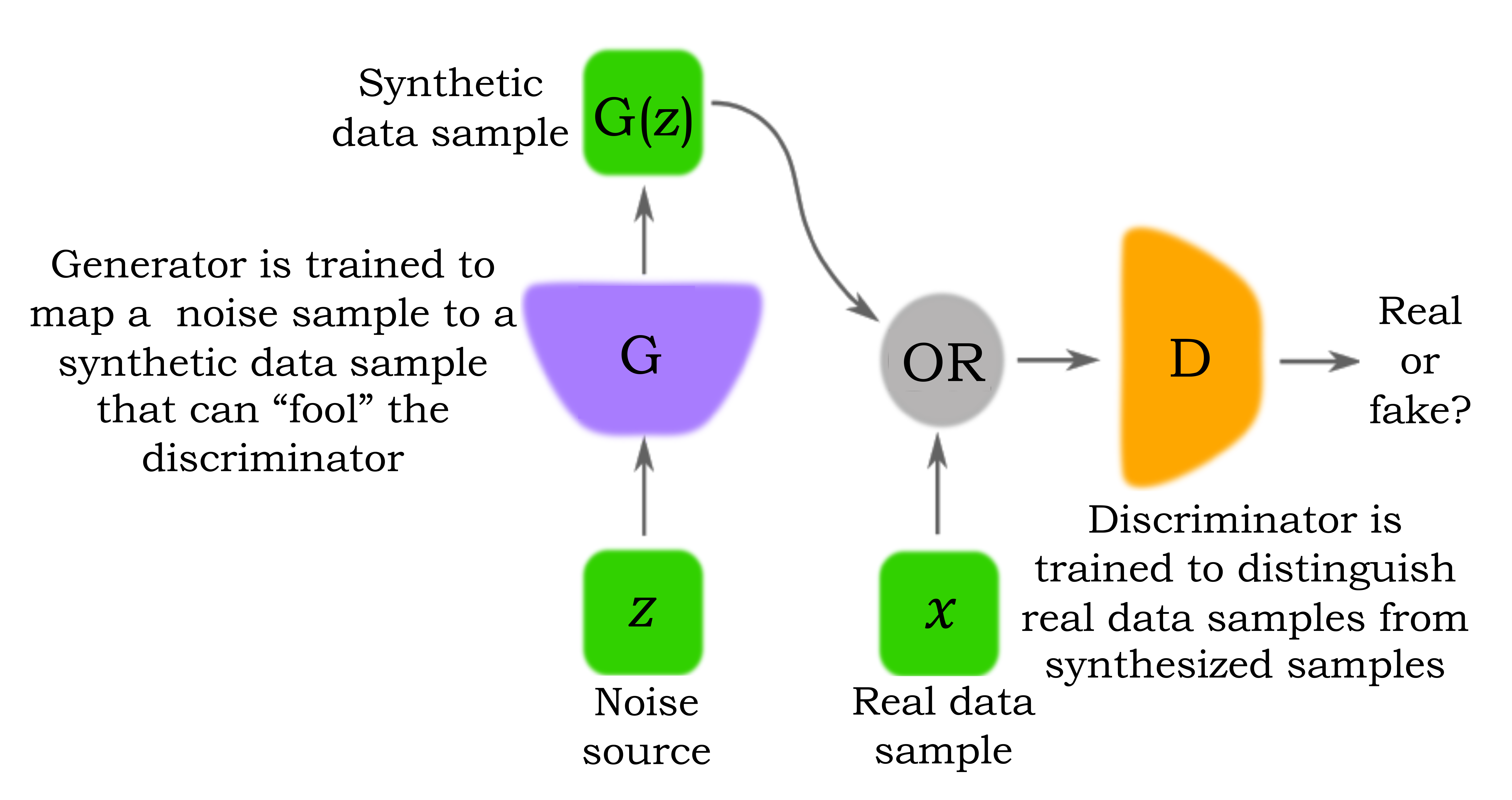}
			%\caption{Generative adversarial net;}
		\end{minipage}	
	}
	\subfloat[cGAN;]
	{
		\begin{minipage}[t]{0.48\textwidth}
			\centering
			\includegraphics[width=0.7\textwidth,height = 3.8cm]{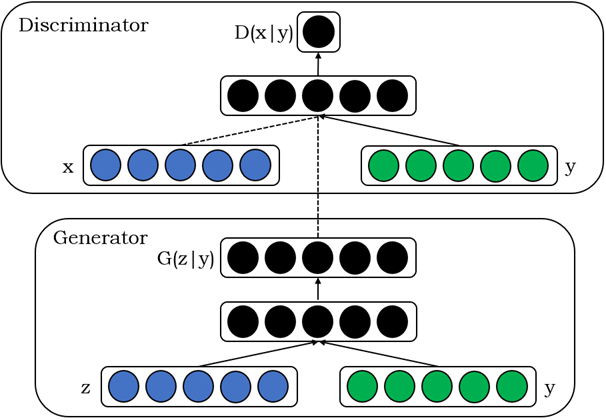}
			%\caption{Conditional generative adversarial net.}
		\end{minipage}
	}
	
	\subfloat[CycleGAN.]
	{
		\begin{minipage}[t]{1\textwidth}
			\centering
			\includegraphics[width=0.8\textwidth, height = 3cm]{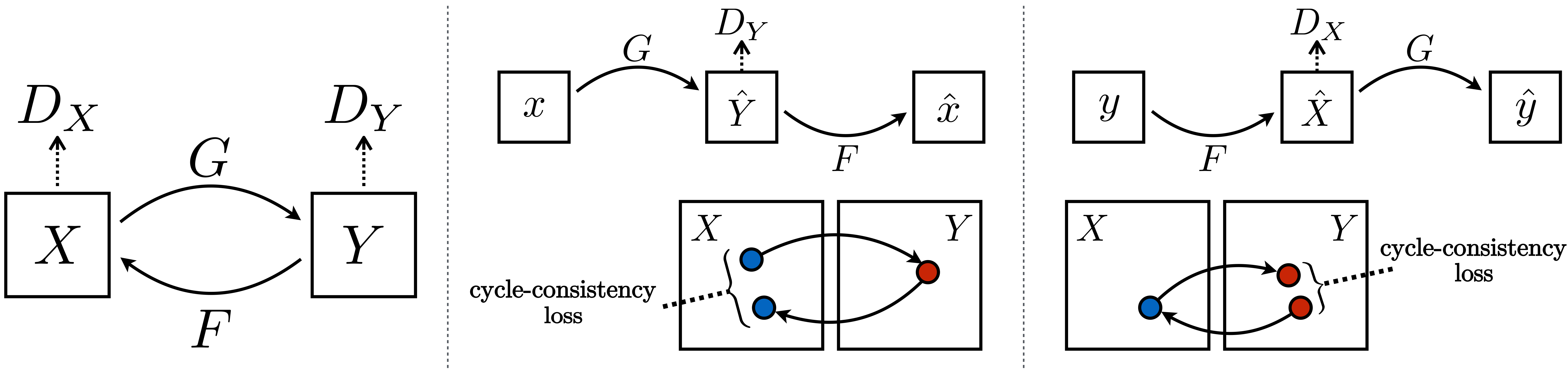}
			%\caption{Generative adversarial net;}
		\end{minipage}	
	}
	\caption{Generative adversarial networks and several typical variants. (a). Generative adversarial networks \copyright (2018) IEEE. Reprinted, with permission, from \cite{creswell2018generative}; (b). Conditional generative adversarial networks \cite{mirza2014conditional}; (c). Cycle-Consistent adversarial networks \copyright (2017) IEEE. Reprinted, with permission, from \cite{zhu2017unpaired}.}
	\label{GAN_fig}
\end{figure*}

\textbf{Domain adaptation.} 
There are many machine learning techniques that are connected to transfer learning \cite{pan2009survey}, for example, domain adaptation \cite{pan2010domain} related to transductive transfer learning; multi-task learning \cite{evgeniou2004regularized} and self-taught learning \cite{raina2007self} related to inductive transfer learning, etc. 
In the review, we put emphasis on domain adaptation where the source and target domains share the same feature spaces, while the feature distributions are different but related. The difference between domain adaptation and transductive transfer learning is that domain adaptation leverages labeled data in the source domain to learn a classifier for the target domain, where the target domain is either fully unlabeled (unsupervised domain adaptation) or has few labeled samples (semi-supervised domain adaptation) \cite{ganin2014unsupervised}. Domain adaptation is promising for the transferability of perception tasks of autonomous systems, because it is efficient to reduce the domain shift among different datasets, arising from synthetic and real images \cite{hoffman2016fcns}, different weather conditions \cite{wulfmeier2017addressing}, different lighting conditions \cite{bak2018domain}, and different seasons \cite{hoffman2017cycada}, etc. Domain adaptation for visual applications includes shallow and deep methods \cite{csurka2017domain}. There are some results studying shallow domain adaptive methods, which mainly include homogeneous domain adaptation and heterogeneous domain adaptation, according to whether the source data and target data have the same representation \cite{pan2010domain}, \cite{zhu2011heterogeneous}, \cite{yang2015learning}, etc. Readers who want to learn more about shallow domain adaptation methods are referred to \cite{csurka2017domain}, \cite{patel2015visual} and the references therein. In this review, we mainly focus on deep domain adaptation methods, including traditional DL \cite{pan2010domain}, \cite{chen2012marginalized}, \cite{long2015learning} and adversarial learning \cite{tzeng2017adversarial}, \cite{ganin2016domain}, \cite{chen2018re}.

%In computer vision tasks, the accuracy and transferability usually conflicts with each other \cite{tan2018survey}, \cite{yosinski2014transferable}, which means that when a well-trained model is transferred to other scenarios, its accuracy may be strongly reduced. Various domain adaptive algorithms have been proposed for transferring learning problem: including supervised \cite{daume2009frustratingly}, semi-supervised \cite{kumar2010co} and unsupervised DA \cite{ganin2014unsupervised}, \cite{hu2018duplex}. Additionally, GANs are typical methods in adversarial learning. By utilizing GANs as the domain adaptive method, the discriminator can ensure that the network cannot distinguish between training or test domain distribution \cite{tzeng2017adversarial}. In other words, adversarial learning via GANs is promising to limit domain shift and can achieve better domain adaptation.

\subsection{Adversarial learning}
Early adversarial learning modeled the learner and the adversary as a competitive two-player game \cite{dalvi2004adversarial}. Subsequently, adversarial learning games are expanded into different forms, including a Bayesian game \cite{grosshans2013bayesian}, a sequential  game \cite{bruckner2012static}, and a bi-level optimization problem \cite{mei2015using}, etc \cite{dasgupta2020playing}. With the popularity of DNNs, Goodfellow \textit{et al.} \cite{goodfellow2014generative} used adversarial learning for generating tasks, i.e., generative adversarial networks (GANs). This model is widely used in various fields of autonomous systems.

\textbf{Generative adversarial networks.} As a powerful learning-based method for computer vision tasks, adversarial learning not only improves the accuracy, but also helps improve the transferability of the model by reducing the differences between the training and testing domain distributions \cite{tzeng2017adversarial}. GANs are architectures that use adversarial learning methods for generative tasks \cite{gui2020review}. The framework includes two models, a generator $G$ and a discriminator $D$, as shown in Fig. \ref{GAN_fig}. $G$ captures the prior noise distribution $p_z(z)$ to generate fake data $G(z)$, and $D$ outputs a single scalar to characterize whether the sample comes from training data $x$ or generated data $G(z)$. $G$ and $D$ play against each other, promote each other, and finally reach the Nash equilibrium \cite{goodfellow2016nips}. $G$ and $D$ focus on a two-player minimax game with the value function $V (G, D)$:
\begin{equation}\label{gan_function}
\begin{aligned}
\min\limits_{G} \max\limits_{D}{V}(G, D) = &  \mathbb{E}_{x\sim p_{\scriptsize \rm data}(x)}[\log D(x)]\\ & + \mathbb{E}_{z\sim p_z(z)}[\log(1-D(G(z)))],
\end{aligned}
\end{equation}
where ${V}(G,D)$ is a binary cross-entropy function, which aims to let $D$ classify real or fake samples. In Eq. (\ref{gan_function}), $D$ tries to maximize its output, $G$ tries to minimize its output, and the game ends at a saddle point\cite{goodfellow2016nips}.

\textbf{Conditional generative adversarial networks.} In the original generative model, since the prior comes from the noise distribution $p_z(z)$, the mode of the generated data cannot be controlled \cite{goodfellow2016nips}. Mirza \textit{et al.} \cite{mirza2014conditional} then proposed conditional generative adversarial networks (cGANs), in which some extra information $y$ is fed to the generator and discriminator in the model, such that the data generation process can be guided, as shown in Fig. \ref{GAN_fig}. Note that $y$ can be class labels or any other kind of auxiliary information. Compared with Eq. (\ref{gan_function}), the objective function of cGANs is as follows:
\begin{equation}\label{cgan_function}
\begin{aligned}
\min\limits_{G} \max\limits_{D}{V}(G,D) = &  \mathbb{E}_{x\sim p_{\scriptsize \rm data}(x)}[\log D(x | y)]\\ & + \mathbb{E}_{z\sim p_{\scriptsize \rm z}(z)}[\log(1- D(G(z | y)))].
\end{aligned}
\end{equation}

\textbf{Cycle-consistent adversarial networks.} Unlike models tailored for specific tasks, like GANs and cGANs, cycle-consistent adversarial networks (CycleGAN) use a unified framework for various image tasks, which make the framework simple and effective \cite{zhu2017unpaired}. Zhu \textit{et al.} \cite{zhu2017unpaired} proposed CycleGAN to learn image translation between the source domain $X$ and the target domain $Y$ with unpaired training examples ${\{x_i\}}^N_{i=1} \in X$ and ${\{y_j\}}^M_{j=1} \in Y$, in which $N$, $M$ represent the total number of samples in the source and target domains, as shown in Fig. \ref{GAN_fig}. The framework includes two generators $G: X \rightarrow Y$ and $F: Y \rightarrow X$, and two discriminators $D_X$ and $D_Y$, where $D_X$ distinguishes between images $x$ and translated images $F(y)$, similarly, $D_Y$ distinguishes between images $y$ and translated images $G(x)$. The output of the mapping $G$ is $\hat{y} = G(x)$, and the output of the mapping $F$ is $\hat{x} = F(y)$. Zhu \textit{et al.} express the adversarial loss for the generator $G: X \rightarrow Y$ and the discriminator $D_Y$ as follows:
\begin{equation}\label{cyclegan_ad}
\begin{aligned}
\mathcal{L}_{GAN}(G,D_Y,X,Y) = &\mathbb{E}_{y \sim p_{\scriptsize \rm data}(y)}[\log D_Y(y)]\\ &+ \mathbb{E}_{x \sim p_{\scriptsize \rm data}(x)}[\log (1-D_Y(G(x))].
\end{aligned}
\end{equation}
They similarly define the adversarial loss for the generator $F: Y \rightarrow X$ and the discriminator $D_X$ as $\mathcal{L}_{GAN}(F,D_X,Y,X)$.
Based on the adversarial loss, they proposed a cycle consistency loss to encourage $F(G(x)) \approx x$ and $G(F(y)) \approx y$. The cycle consistency loss is expressed as:
\begin{equation}\label{cycleGAN_cyc}
\begin{aligned}
\mathcal{L}_{cyc}(G,F) = &\mathbb{E}_{x \sim p_{\scriptsize \rm data}(x)}[\left\|F(G(x))-x\right\|_1]\\ &+ \mathbb{E}_{y \sim p_{\scriptsize \rm data}(y)}[\left\|G(F(y))-y\right\|_1].
\end{aligned}
\end{equation}
The full objective of CycleGAN is:
\begin{equation}\label{cyclegan}
\begin{aligned}
\min\limits_{G,F} \max\limits_{D_X,D_Y}\mathcal{L}(G,F,D_X,D_Y) = &\mathcal{L}_{GAN}(G,D_Y,X,Y)\\& + \mathcal{L}_{GAN}(F,D_X,Y,X)\\& + \lambda \mathcal{L}_{cyc}(G,F),
\end{aligned}
\end{equation}
where $\lambda$ is a hyperparameter used to control the relative importance of the adversarial loss and the cycle consistency loss.

As a powerful generative model, many variants of GANs were presented by modifying loss functions or network architectures and they were used for various computer vision tasks. In this review, we mainly focus on the problem of scene transfer and task transfer in autonomous systems using GANs, including image style transfer, image SR, image denoising/dehazing/rain removal, semantic segmentation, depth estimation, pedestrian detection and person re-ID.

\subsection{Reinforcement learning}
Reinforcement learning (RL) is the problem faced by an agent that learns behavior through trial-and-error interactions in a dynamic environment \cite{kaelbling1996reinforcement}. In the RL framework, an agent interacts with the environment to choose the action in the state of a given environment in order to maximize its long-term reward \cite{sutton2018reinforcement}. 
%RL methods have been applied to a wide range of controlling tasks, from locomotion \cite{kohl2004policy} to manipulation \cite{peters2008reinforcement}. 
RL algorithms can be classified into two kinds, model-based and model-free algorithms \cite{geffner2018model}. Model-based RL is to learn a transition model that allows the environment to be simulated without directly interacting with the environment \cite{arulkumaran2017brief}. Model-based methods include guided policy search (GPS) \cite{levine2016end}, and model-based value expansion (MBVE) \cite{feinberg2018model}, etc. However, model-free RL uses the experience of states and environments directly to generate actions \cite{glascher2010states}. Model-free methods include deep Q network (DQN) \cite{mnih2013playing}, deep deterministic policy gradient (DDPG) method \cite{lillicrap2015continuous}, dynamic policy programming (DPP) method \cite{azar2012dynamic} and asynchronous advantage actor-critic (A3C) method \cite{zhu2017target}, etc. Model-free algorithms can learn complex tasks but tend to be inefficient in sampling, while model-based algorithms are more efficient in sampling, but usually have difficulty in scaling to complicated tasks \cite{kahn2018self}. With further research on the application of RL methods, several problems occur that model-based algorithms are no longer applicable to more complex tasks, while model-free algorithms need more training data. Moreover, when the given environment changes or training data are insufficient, chances are that RL methods need to train the model starting from the scratch, which is inefficient and inaccurate. Therefore, RL methods are limited when generalizing to different tasks and domains \cite{kahn2018self}. In this review, we mainly focus on several modifications on RL methods, such as amending the network structure \cite{zhang2016modular}, \cite{polvara2017autonomous} and optimizing the way of training \cite{mirowski2016learning}, \cite{zeng2018learning}, in order to enable the model to learn the new tasks accurately in the same domain or transferably across domains.

% As shown in recent work, this can be difficult in domains with severe discontinuities in the dynamics and reward function.

% Since RL methods use relatively low-dimensional policy representations due to reward sparsities, high-dimensionality of sensorimotor spaces and  which are inherent in such problems \cite{chen2018deep}. 

%Therefore, combining RL with deep neural network is a way to improve these performance. DRL can train highly flexible and versatile deep neural networks to obtain action selection strategies for complex problems \cite{chen2018deep}.

\subsection{Meta-learning}
Meta-learning, or ``learning to learn", ``learn how to learn", i.e., using previous knowledge and experience to guide the learning of new tasks in order to equip the model with the ability to learn across domains \cite{vanschoren2018meta}. The goal of meta-learning is to train a model that can quickly adapt to a new task using only a few data points and training iterations \cite{finn2017model}. Similar to transfer learning, meta-learning improves the learner's generalization ability in multi-task setting. However, unlike transfer learning, meta-learning focuses on the sampling of both data and tasks. Therefore, meta-learning models are trained by being exposed to a large number of tasks, which qualifies them to learn new tasks from few data settings. The meta-learning methods can be divided into three categories: recurrent models, metric learning and learning optimizers \cite{li2017meta}. 

Recurrent models are trained by various methods, such as long short-term memory (LSTM) \cite{hochreiter1997long} and temporal convolution network (TCN) \cite{lea2017temporal}, to acquire the dataset sequentially and then process new inputs from the task. LSTM \cite{hochreiter1997long} processes data sequentially and figures out its own learning strategy from the scratch. Moreover, TCN \cite{lea2017temporal} uses convolution structures to capture long-range temporal patterns, whose framework is simpler and more accurate than LSTM.

Metric learning is a way to calculate the similarity between two targets from different tasks. For a specific task, the input target is classified into a target category with large similarity judging from a metric distance function \cite{kulis2013metric}. It has been widely used for few-shot learning \cite{snell2017prototypical}, during which, the data belong to a large number of categories, some categories are unknown at the stage of training and the training samples of each category are particularly small \cite{wang2019generalizing}. These characteristics are consistent with the characteristics of meta-learning. There are four sorts of typical networks proposed for metric learning, siamese network \cite{chopra2005learning}, prototypical network \cite{li2017meta}, matching network \cite{vinyals2016matching} and relation network\cite{sung2018learning}. 

\begin{figure}[h]
\centering
\includegraphics[width=9cm,height=4cm]{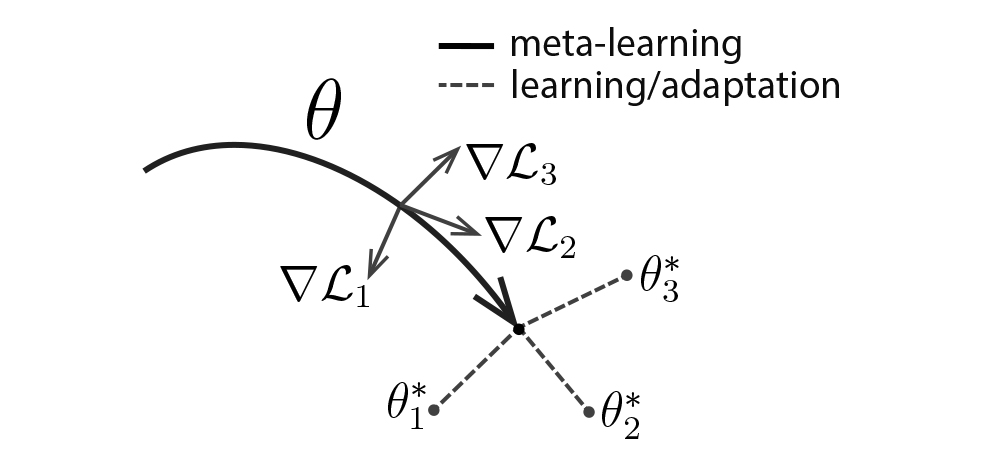}
\caption{Diagram of the MAML algorithm, which optimizes for a representation $\theta$ that can quickly adapt to new tasks \cite{finn2017model}.}
\label{fig_MAML}
\end{figure}

Learning an optimizer, that is, one meta-learner learns how to update the learner so that the learner can learn the task efficiently \cite{lee2018gradient}. This method has been extensively studied to obtain better optimization results of neural networks. Combined with RL \cite{rakelly2019efficient} or imitation learning \cite{duan2017one}, meta-learning is able to learn new policies accurately or adapt to new tasks effectively. Model-agnostic meta-learning (MAML) \cite{finn2017model} is a representative and popular meta-learning optimization method, which uses stochastic gradient descent (SGD) \cite{bottou2010large} to update. It adapts quickly to new tasks due to no assumptions made about the form of the model and no extra parameters introduced for meta-learning. MAML includes a base-model learner and a meta-learner. Each base-model learner learns a specific task and the meta-learner learns the average performance $\theta$ of multiple specific tasks as the initialization parameters of one new task \cite{finn2017model}.
From Fig. \ref{fig_MAML}, the model is represented by a parametrized function
$f_{\theta}$ with the parameter $\theta.$ When adapting to a new task $\mathcal{T}_{i}$ that is drawn from a distribution over tasks $p(\mathcal{T}),$ the model's parameter is updated to $\theta_{i}^{\prime}$. $\theta_{i}^{\prime}$ is computed by one or more gradient descent updates on task $\mathcal{T}_{i}.$  Moreover, $\mathcal{L}_{\mathcal{T}i}$ represents the loss function for task  $\mathcal{T}_{i}$ and the step size $\alpha$ is regarded as a hyperparameter. For example, we consider one gradient update on task  $\mathcal{T}_{i},$

\begin{equation}
\theta_{i}^{\prime}=\theta-\alpha \nabla_{\theta} \mathcal{L}_{\mathcal{T}_{i}}\left(f_{\theta}\right).
\end{equation}

The model parameters are trained by optimizing for the performance of a parametrized function $f_{\theta_{i}^{\prime}}$ with parameter $\theta_{i}^{\prime},$ corresponding to the following problem:

\begin{equation}\min _{\theta} \sum_{\mathcal{T}_{i} \sim p(\mathcal{T})} \mathcal{L}_{\mathcal{T}_{i}}(f_{\theta_{i}^{\prime}})=\sum_{\mathcal{T}_{i} \sim p(\mathcal{T})} \mathcal{L}_{\mathcal{T}_{i}}(f_{\theta-\alpha \nabla_{\theta} \mathcal{L}_{\mathcal{T}_{i}}(f_{\theta})}).\end{equation}

When extending MAML to the imitation learning setting, the model's input, $\mathbf{o}_{t}$, is the agent's observation sampled at time $t,$ whereas the output $\mathbf{a}_{t}$ is the agent's action taken at time $t.$ The demonstration trajectory can be represented as $\tau:=\left\{\mathbf{o}_{1}, \mathbf{a}_{1}, \ldots \mathbf{o}_{T}, \mathbf{a}_{T}\right\}$, using a mean squared error loss as a function of policy parameters $\phi$ as follows:
\begin{equation}\mathcal{L}_{\mathcal{T}_{i}}\left(f_{\phi}\right)=\sum_{\tau^{(j)} \sim \mathcal{T}_{i}} \sum_{t}\left\|f_{\phi}\left(\mathbf{o}_{t}^{(j)}\right)-\mathbf{a}_{t}^{(j)}\right\|_{2}^{2}.\end{equation}

During meta-training, several demonstrations are sampled as training tasks. The demonstrations help to compute $\theta_{i}^{\prime}$ for each task $\mathcal{T}_{i}$ using gradient descent with Eq. (6) and to compute the gradient of the meta-objective by using Eq. (7) with the loss in Eq. (8). During meta-testing, we consider using only a single demonstration as a new task $\mathcal{T}$, updating with SGD. Therefore, the model is updated to acquire a policy for that task \cite{finn2017one}.

\subsection{The relationship between adversarial learning, RL and meta-learning}

RL \cite{sutton1998introduction} is a method to describe and solve the problem that agents learn policies to achieve the maximum returns or specific goals in the interactions with the environment. Pfau \textit{et al.} \cite{pfau2016connecting} discussed the connection between GANs and actor-critic (AC) methods. AC is a kind of RL method that learns the policy and value function simultaneously. To be specific, the actor network chooses the proper action in a continuous action space, while the critic network implements a single-step-update, which improves the learning efficiency \cite{konda2000actor}. Pfau \textit{et al.} argued that GANs can be viewed as an AC approach in an environment where actors cannot influence rewards. RL and GANs are integrated for various tasks, such as real-time point cloud shape completion \cite{sarmad2019rl} and image synthesis \cite{ganin2018synthesizing}, etc. 

In the field of RL, using the cost function to understand the underlying behavior is called inverse reinforcement learning (IRL) \cite{ng2000algorithms}. The policy distribution in the IRL can be regarded as the data distribution of the generator in GANs, and the reward in the IRL can be regarded as the discriminator in GANs. However, IRL learns the cost function to explain expert behavior, but cannot directly tell the learner how to take action, which leads to high running costs. Ho \textit{et al.} \cite{ho2016generative} proposed generative adversarial imitation learning (GAIL), combining GANs with imitation learning, which employs GANs to fit the states and actions distributions that define expert behavior. GAIL significantly improves the performance in large-scale and high-dimensional planning problems \cite{li2017infogail}.

Introducing meta-learning to RL methods is called meta-RL methods \cite{hochreiter2001learning}, which equips the model to solve new problems more efficiently by utilizing the experience from prior tasks. A meta-RL model is trained over a distribution of different but related tasks, and during testing, it is able to learn to solve a new task quickly by developing a new RL algorithm \cite{wang2016learning}. There are several meta-RL algorithms that utilize the past experience to achieve good performance on new tasks. For example, MAML \cite{finn2017model} and Reptile \cite{nichol2018reptile} are typical methods on updating model parameters and optimizing model weights; MAESN (model agnostic exploration with structured noise) \cite{gupta2018meta} can learn structured action noise from prior experience; EPG (evolved policy gradient) \cite{houthooft2018evolved} defines the policy gradient loss function as a temporal convolution over the past experience. Moreover, when dealing with unlabeled training data, unsupervised meta-RL methods \cite{gupta2018unsupervised} effectively acquire accelerated RL procedures without manual task design, such as collecting data and labeling data, etc. Therefore, both supervised and unsupervised meta-RL can transfer previous tasks information to new tasks across domains.

\section{Autonomous Systems Meet \\Accuracy \& Transferability} \label{3}

Computer vision and robot control tasks are critical to autonomous systems. Currently, there are a variety of learning-based methods for perception and decision-making tasks. As mentioned at the beginning of Section \ref{2}, most traditional learning-based methods show good accuracy in the same data distribution or task, but suffer from poor transferability, i.e., when considering the application of a well-trained model to different scenarios, its accuracy often decreases heavily. This is due to the obvious domain gap between different datasets. Therefore, domain adaptation between different domains is very important for autonomous systems. 

\subsection{Overview of the section}
In this section, we mainly focus on learning-based methods in the perception and decision-making tasks of autonomous systems, from the perspectives of accuracy or transferability or both of them, like image style transfer, image super-resolution, image  denoising/dehazing/rain removal, semantic segmentation, depth estimation, other geometry information (surface normal and optical flow) prediction, pedestrian detection/re-ID/tracking, robot navigation and robotic manipulation in autonomous systems. Although some traditional DL-based methods mainly focus on improving the accuracy of the model, in recent years, methods for the above visual tasks have gradually attached importance to the transferability, using adversarial learning, RL and meta-learning, etc. We summarize some typical computer vision tasks and robot control tasks in autonomous systems in Table \ref{table3} and Table \ref{table4}, including their training manners, loss functions, learning methods and experimental platforms, etc. As shown in Table \ref{table3}, the training manners of some computer vision tasks gradually change from supervised to unsupervised ones, and their loss functions change from accuracy to transferability between domains. Table \ref{table4} in subsection \ref{H} indicates that, as for robot control tasks, informative simulation environments and flexible practice platforms will help to accurately transfer information across domains.

\newcommand{\tabincell}[2]{\begin{tabular}{@{}#1@{}}#2\end{tabular}}

\begin{table*}
	\centering
	\begin{threeparttable}
	\caption{Summary of methods for computer visual tasks in autonomous systems}
	\begin{tabular}{cccccccc}
		\toprule
		Year & Reference & Task & Multi-task & GANs-based & Supervision\tnote{1} & Loss\tnote{2}\\
		\midrule
		2016 & Gatys \textit{et al.} \cite{gatys2016image} & Style transfer &   &   & Supervised & C \\
		2016 & Johnson \textit{et al.} \cite{johnson2016perceptual} & Style transfer  & $\surd$ &  & Supervised & B\\
		2017 & Li \textit{et al.} \cite{li2017demystifying} & Style transfer &   &   & Supervised & C\\
		2017 & Pix2Pix \cite{isola2017image} & Style transfer & $\surd$ & $\surd$ & Supervised & A, E \\
		2017 & CycleGAN \cite{zhu2017unpaired} & Style transfer & $\surd$ & $\surd$ & Unsupervised & A, D \\
		2019 & DLOW \cite{gong2019dlow} & Style transfer & $\surd$ & $\surd$ & Unsupervised & A, D\\
		2019 & INIT \cite{shen2019towards} & Style transfer &  & $\surd$ & Unsupervised & A, C \\
		\midrule
		2014 & SRCNN \cite{dong2014learning} & Super-resolution & & & Supervised & F \\
		2015 & SRCNN \cite{dong2015image} & Super-resolution & & & Supervised & F \\
		2016 & FSRCNN \cite{dong2016accelerating} & Super-resolution & & & Supervised & F \\
		2016 & Johnson \textit{et al.} \cite{johnson2016perceptual} & Super-resolution  & $\surd$ &  & Supervised & B  \\
		2017 & SRGAN \cite{ledig2017photo} & Super-resolution & & $\surd$ & Supervised & A, F \\
		2017 & EnhanceNet \cite{sajjadi2017enhancenet} & Super-resolution & & $\surd$ & Supervised & A, B, F \\
		2018 & ZSSR \cite{shocher2018zero} & Super-resolution & & & Unsupervised & E \\
		2018 & ESRGAN \cite{wang2018esrgan} & Super-resolution & & $\surd$ & Supervised & A, B, E \\
		2018 & CinCGAN \cite{yuan2018unsupervised} & Super-resolution & & $\surd$ & Unsupervised & A, D, F \\
		2019 & Soh \textit{et al.} \cite{soh2019natural} & Super-resolution & & $\surd$ & Supervised & A, C, F\\
		%2020 & LSRGANs \cite{zhong2020optimizing} & super resolution & & $\surd$ & supervised & GAN loss, Cosine Contextual loss\\
		2020 & Gong \textit{et al.} \cite{gong2020learning} & Super-resolution & & $\surd$ & Unsupervised & A, D, E\\
		\midrule
		2018 & DeblurGAN \cite{kupyn2018deblurgan} & Image deblurring & & $\surd$ & Supervised & A, B \\
		2019 & DeblurGAN-v2 \cite{kupyn2019deblurgan} & Image deblurring & & $\surd$ & Supervised & A, E, F \\
		2019 & Dr-Net \cite{aljadaany2019douglas} & Image deblurring & &$\surd$ & Supervised & A, E\\
		2018 & Li \textit{et al.} \cite{li2018single} & Image dehazing & & $\surd$ & Supervised & A, B, E \\
		2018 & Cycle-Dehaze \cite{engin2018cycle} & Image dehazing & & $\surd$ & Unsupervised & A, D \\
		2019 & Kim \textit{et al.} \cite{kim2019bidirectional} & Image dehazing & & $\surd$ & Supervised & A, D, E, F\\
		2019 & CDNet \cite{dudhane2019cdnet} & Image dehazing & & $\surd$ & Unsupervised & A, D\\
		2020 & Sharma \textit{et al.} \cite{sharma2020scale} & Image dehazing &  & $\surd$ & Supervised & A, B, E, F\\
		2018 & Qian \textit{et al.} \cite{qian2018attentive} & Image rain removal & & $\surd$ & Supervised & A, B, F \\
		2019 & Li \textit{et al.} \cite{li2019heavy} & Image rain removal & & $\surd$ & Supervised & A, B, F\\
		2019 & ID-CGAN \cite{zhang2019image} & Image rain removal & & $\surd$ & Supervised & A, B, E\\
		2020 & AI-GAN \cite{jin2020ai} & Image rain removal & & $\surd$ & Supervised & A, F\\
		\midrule
		2016 & Hoffman \textit{et al.} \cite{hoffman2016fcns} & Semantic segmentation &   &  & Unsupervised & F, G\\
		2017 & SegNet \cite{badrinarayanan2017segnet}  &  Semantic segmentation &   &  & Supervised & F \\
		2017 & Mask R-CNN \cite{he2017mask} & Instance segmentation &   & & Supervised & F \\
		2017 & CyCADA \cite{hoffman2017cycada} & Semantic segmentation &   & $\surd$ & Unsupervised & A, D, F\\
		2018 & FCAN \cite{zhang2018fully} & Semantic segmentation &   &$\surd$ & Unsupervised & A, F\\
		2018 & Hu \textit{et al.} \cite{hu2018learning} & Instance segmentation &   & & Partially supervised\tnote{3} & F\\
		2018 & Hong \textit{et al.} \cite{hong2018conditional} & Semantic segmentation &   & $\surd$ & Unsupervised & A, F, G\\
		2019 & CrDoCo \cite{chen2019crdoco} & Semantic segmentation & $\surd$ & $\surd$ & Unsupervised & A, C, D, F\\
		2019 & CLAN \cite{luo2019taking} & Semantic segmentation &   & $\surd$ & Unsupervised & A, F \\
		2019 & Li \textit{et al.} \cite{li2019bidirectional} & Semantic segmentation &   & $\surd$ & Self-supervised & A, B, C, F \\
		2020 & Erkent \textit{et al.} \cite{erkent2020semantic} & Semantic segmentation &   & $\surd$ & Unsupervised & A, F \\
		\midrule
		2014 & Eigen \textit{et al.} \cite{eigen2014depth} & Depth estimation&    &  & Supervised & F\\
		2015 & Eigen \textit{et al.} \cite{eigen2015predicting} & Depth estimation & $\surd$ & & Supervised & F\\
		2015 & Liu \textit{et al.} \cite{liu2015learning} & Depth estimation &   & & Supervised & F\\
		2018 & Atapour-Abarghouei \textit{et al.} \cite{atapour2018real} & Depth estimation & & $\surd$ & Supervised & A, C\\
		2019 & ASM \cite{hwang2019adversarial} & Depth estimation & $\surd$ &  & Supervised & F\\
		%2019 & Chen \textit{et al.} \cite{chen2019towards} & \tabincell{c}{semantic segmentation\\ depth prediction} & & unsupervised & \tabincell{c}{reconstruction loss\\left-right consistency loss \\ disparity smoothness loss\\semantic segmentation loss}\\
		2019 & CrDoCo \cite{chen2019crdoco} & Depth estimation & $\surd$ & $\surd$ & Unsupervised & A, C, D, F\\
		2019 & GASDA \cite{zhao2019geometry} & Depth estimation & & $\surd$ & Unsupervised & A, D, F \\
		2020 & ARC \cite{zhao2020domain} & Depth estimation & & $\surd$ & Supervised & A, B, C, D, F\\
		\midrule
		2013 & ConvNet \cite{sermanet2013pedestrian}  & Pedestrian detection & & & Unsupervised & F\\
		%2013 & Ouyang \textit{et al.} \cite{2013joint} & Pedestrian detection & & & Supervised & F\\
		2015 & TA-CNN \cite{tian2015pedestrian} & Pedestrian detection & & & Supervised & F\\
		2017 & SAF R-CNN \cite{li2017scale} & Pedestrian detection & & & Supervised & F\\
		2019 & Kim \textit{et al.} \cite{kim2019unpaired} & Pedestrian detection & & $\surd$ & Unsupervised & A, E, F\\
		2018 & SPGAN \cite{deng2018image} & Person re-ID & & $\surd$ & Unsupervised & A, D, F \\
		2018 & CamStyle \cite{zhong2018camera} & Person re-ID & & $\surd$ & Unsupervised & A, D, F\\
		2019 & ATNet \cite{liu2019adaptive} & Person re-ID & & $\surd$ & Unsupervised & A, D, F\\
	    2017 &  ADNet \cite{yun2017action} &  Pedestrian tracking &  $\surd$ & &  Supervised &  F\\ 
	    2017 &  Supancic \textit{et al.} \cite{supancic2017tracking}  &  Pedestrian tracking & & &  Supervised &  /\tnote{4}\\
	    2018 &  Chen \textit{et al.}  \cite{chen2018real}  &  Pedestrian tracking & & &  Supervised &  E \\
		2019 &  ConvNet-LSTM \cite{luo2019end}  &  Pedestrian tracking & & &  Supervised &  /\tnote{4}\\

		\bottomrule
	\end{tabular}
	\label{table3}
	\begin{tablenotes}
        \footnotesize
        \item[1] For models that do not explicitly state whether they are supervised in the references, this review considers models that require paired images as supervised and models that do not require paired images as unsupervised.
        \item[2] We classify the loss function into several classes. ``A" represents adversarial (GAN) loss. ``B" represents perceptual loss. ``C" represents reconstruction loss. ``D" represents cycle consistency loss. ``E" represents pixel-wise loss. ``F" represents specific task loss like depth loss and semantic loss, etc. ``G" represents domain transfer loss like domain adversarial loss and domain classifier loss, etc.
       \item[3] Partially supervised learning problems refer to training on the combination of strong and weak labels \cite{hu2018learning}. According to \cite{schwenker2014pattern}, partially supervised learning include: active learning, general semi-supervised learning, semi-supervised learning with graphs, partially supervised learning in ensembles and multiple classifier systems.
        \item[4] RL-based methods mainly focus on reward and action instead of loss.
      \end{tablenotes}
    \end{threeparttable}
\end{table*}

\subsection{Image style transfer}

\begin{figure*}[htbp]
	\centering
	\subfloat[Results of instance-level day$\rightarrow$night translation. The first row shows the input images, and the second row shows output images;]
	{
		\begin{minipage}[t]{0.7\textwidth}
			\centering
			\includegraphics[width=1\textwidth, height = 3.5cm]{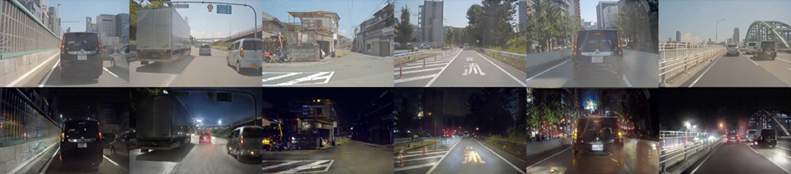}
		\end{minipage}	
	}
	\subfloat[Results of seasonal conversion.]
	{
		\begin{minipage}[t]{0.25\textwidth}
			\centering
			\includegraphics[width=1\textwidth, height = 3.5cm]{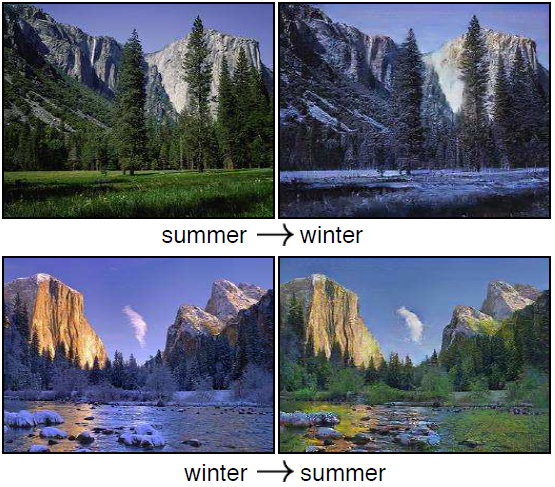}
		\end{minipage}
	}
	\caption{Generative adversarial networks for image style transfer. (a). Results of instance-level day$\rightarrow$night translation \copyright (2019) IEEE. Reprinted, with permission, from \cite{shen2019towards}; (b). Results of seasonal conversion \copyright (2017) IEEE. Reprinted, with permission, from \cite{zhu2017unpaired}. See also Fig. \ref{style transfer} and Table \ref{table3}.}
	\label{style transfer}
\end{figure*}

Images can be well transferred between different styles, which is conducive to the perception and decision-making algorithms of autonomous systems applicable to various scenarios. For autonomous systems, it is inevitably to face the problem of the image style transfer arising from seasonal conversion \cite{hoffman2017cycada}, varying weather conditions \cite{wulfmeier2017addressing} or day conversion \cite{bak2018domain}. In particular, it is more challenging and interesting to consider transferring training data for night to day, rainy to sunny, or winter to summer, since most autonomous systems have a better ability to perceive under good lighting or weather condition than some harsh environments. The task of image style transfer is to change the content of the source domain image to the target domain one, while ensuring that the style is consistent with the target domain \cite{gatys2016image}. In addition, style transfer, as an interesting data augmentation strategy, can extend the range of lighting and weather changes, thus further improving the transferability of the model \cite{shorten2019survey}. As well, using the image style transfer algorithm to achieve the transfer from the simulated environment to the real-world is very useful for semantic segmentation, robot navigation and grasping tasks, because training directly in real-world may lead to higher experimental costs due to some possible damages to hardwares \cite{shorten2019survey}. Traditional methods to achieve style transfer mainly rely on non-parametric techniques to manipulate the pixels of the image like \cite{efros2001image}, \cite{hertzmann2001image}, etc. Although traditional methods have achieved good results in style transfer, they are limited to using only low-level features of the image for texture transfer, but not semantic transfer \cite{gatys2016image}.

\textbf{Traditional DL-based style transfer.} Convolutional neural networks (CNNs) have been used in image style transfer, since they have achieved  fantastic results in numerous visual perception areas. Gatys \textit{et al.} \cite{gatys2016image} first proposed to utilize CNNs (pre-trained VGG-Networks) to separate content and style from natural images, and then combined the content of one image with the style of another into a new image to achieve an artistic style transfer. This work opened up a new viewpoint for style transfer using DNNs. To reduce the computational burden, Johnson \textit{et al.} \cite{johnson2016perceptual} proposed to use the perceptual loss instead of the per-pixel loss for the image style transfer task. This method achieves similar results to \cite{gatys2016image}, while it is three orders of magnitude faster. Since Gatys \textit{et al.} \cite{gatys2016image} used the Gram matrices to represent the artistic style of a image, the subsequent improvement works did not investigate its principles in depth. Li \textit{et al.} \cite{li2017demystifying} first regarded neural style transfer as a domain adaptation problem, and theoretically show that the second order interaction in the Gram matrix is not necessary for style transfer, which is equivalent to a specific maximum mean discrepancy (MMD). In addition, Chen \textit{et al.} \cite{chen2018stereoscopic} presented a stereo neural style conversion that can be used in emerging technologies such as 3D movies or VR. This method seems promising for improving the perception accuracy of autonomous systems in unmanned scenes, because the transferred results contain more stereo information in the scene.

\textbf{GANs-based style transfer.} Traditional CNNs-based methods minimize the Euclidean distance between predicted pixels and ground truth pixels, which may cause blurry results \cite{isola2017image}. GANs can be used for image style transfer, which can produce more realistic images \cite{isola2017image}. Isola \textit{et al.} \cite{isola2017image} used cGANs to image style transfer, and the experimental results showed that cGANs (with L1 loss) not only have satisfactory results for style transfer task, but also can produce reasonable results for a wide variety of problems like semantic segmentation and background removal. However, this method requires paired image samples, which is often difficult to implement in practice. By considering this issue, Zhu \textit{et al.} \cite{zhu2017unpaired} proposed CycleGAN to learn image translation between domains with unpaired examples, as shown in Fig. \ref{style transfer}. As mentioned in Section \ref{2}, the framework of CycleGAN includes two generators and two discriminators to achieve mutual translation between the source and the target domain. The main insight of CycleGAN is to preserve the key attributes between the input and the translated image by using a cycle consistency loss. At almost the same time, DiscoGAN \cite{kim2017learning} and DualGAN \cite{yi2017dualgan} were presented to adopt similar cycle consistency ideas to achieve an image transfer task across domains. In order to improve CycleGAN from the aspect of semantic information alignment at the feature-level, Hoffman \textit{et al.} \cite{hoffman2017cycada} proposed CyCADA by combining domain adaptation and cycle-consistent adversarial, which uniformly considers feature-level and pixel-level adversarial domain adaptation and cycle-consistency constraints. CyCADA has achieved satisfactory results in some challenging tasks, like from synthesis to practical conversion and seasonal conversion, which is very important for the generalization of autonomous systems. It was shown that CyCADA has a better transferability than the original CycleGAN model. Since these methods, like CycleGAN and CyCADA, can only realize the translation between two domains, different models should be trained for each pair of domains in the case of handling multiple domains translation tasks, which limits their wide applications. By considering this point, Choi \textit{et al.} \cite{choi2018stargan} proposed StarGAN to perform image translations for multiple domains using a single generator and a discriminator. StarGAN takes both the image and its domain label as input, and learn to transfer the input image into the corresponding target domain. In order to further improve the existing adaptive image style transfer methods, Gong \textit{et al.} \cite{gong2019dlow} proposed a domain flow generation (DLOW) model, which generates a series of intermediate domains to bridge two different domains. This method may be helpful for gradual changes, such as day or season, because it can generate a continuous sequence of intermediate samples ranging from the source to target samples. Recent image translation tasks focused on semantic consistency of images instead in image style and content. Royer \textit{et al.} \cite{royer2020xgan} proposed XGAN, which is an unsupervised semantic style transfer task for many-to-many mapping. Royer \textit{et al.} used domain adaptation techniques to constrain the shared embedding and proposed a semantic consistency loss as a form of self-supervision to act on two domain translations. This method has a good generalization effect when there is a large domain shift between the two domains. In addition, in order to obtain a fine-grained local information of images, Shen \textit{et al.} \cite{shen2019towards} proposed instance-aware image-to-image translation approach (INIT), which applies instance and global styles to the target image spatially, as shown in Fig. \ref{style transfer}. Similarly, the image style transfer was considered  at the instance level in \cite{ma2018gan}, \cite{mo2018instagan}. 

As a data augmentation strategy, image style transfer can help the scene to be transferred in various lighting conditions, various weather conditions, simulated to real-world environment and so on. Image style transfer helps autonomous systems perform their perception and decision-making tasks in better lighting conditions, and effectively reduces hardware losses in real-world environment, which is critical for autonomous systems. Although many traditional DL-based models have achieved good style transfer results, with the advent of GANs, various research works have been extended based on GANs. The recent developments of image style transfer have focused on instance-level style transfer. We believe that the future works should focus on the image style transfer for more complex scenes, such as changing the style of the specified instance without changing the background style in a wild environment. In addition, some future works should also consider improving the accuracy of style transfer and the speed of the overall process, striving for real-time performance with good accuracy. In addition to the style transfer task, we further consider increasing the resolution of images, i.e., the super-resolution task.

\subsection{Super-resolution}

Super-resolution (SR) is a challenging visual perception task to generate high-resolution (HR) images from low-resolution (LR) image inputs \cite{park2003super}. SR is crucial to understand the environment at high-level for autonomous systems. For example, SR is helpful to construct dense map. In this subsection, we will first discuss the recent developments in SR by focusing on accuracy. Then, we will summarize the new development in SR by considering transferability.

There are a number of methods dedicated to improving image quality, such as single image interpolation \cite{hou1978cubic} and image restoration \cite{zhang2017learning}. It is worth pointing out that they are different from SR. On the one hand, single image interpolation usually cannot restore high-frequency details \cite{hou1978cubic}. In addition, image restoration often uses methods, like image sharpening, in which the input image and output image remain the same size, although the output quality can be improved \cite{zhang2017learning}. SR does not only improve the output quality, but also increases the number of pixels per unit area, i.e., the size of image increases \cite{park2003super}. In some cases, the image SR can be regarded as a method of image enhancement \cite{ding2017towards}. Recently, a large number of SR methods have been proposed, such as interpolation-based methods \cite{keys1981cubic} and reconstruction-based methods \cite{sun2008image}, etc. Farsiu \textit{et al.} \cite{farsiu2004advances} introduced the advances and challenges of traditional methods for SR. 

\textbf{Traditional DL-based SR.} There are some results studying traditional DL-based methods without adversarial learning for SR, which are mainly CNNs-based. Dong \textit{et al.} \cite{dong2014learning} considered using CNNs to handle SR task in an end-to-end manner. They presented the super-resolution convolutional neural network (SRCNN), which has little extra pre/post-processing beyond optimization. In addition, they confirmed that DL provides a better quality and speed for SR than the sparse coding method \cite{yang2010image} and the K-SVD-based method \cite{zeyde2010single}, while SRCNN only uses information on the luminance channel. Dong \textit{et al.} \cite{dong2015image} then extended SRCNN to process three color channels simultaneously to improve the accuracy of SR results. Considering the poor real-time performance of SRCNN, Dong \textit{et al.} \cite{dong2016accelerating} utilized a compact hourglass-shape CNN structure to accelerate the current SRCNN. In fact, most learning-based SR methods use the per-pixel loss between the output image and the ground-truth image \cite{dong2014learning}, \cite{dong2015image}. Johnson \textit{et al.} \cite{johnson2016perceptual} considered the use of perceptual loss to achieve a better SR, which is able to better reconstruct details than the per-pixel loss. Note that the above mentioned SR methods often rely on specific training data. When there are non-ideal imaging conditions due to noise or compression artifacts, the above methods usually fail to provide good SR results. Therefore, Shocher \textit{et al.} \cite{shocher2018zero} considered ``Zero-Shot" SR (ZSSR), which does not rely on prior training. To the best of our knowledge, ZSSR is the first unsupervised CNN-based SR method, which achieves reasonable SR results in some complex or unknown imaging conditions. Due to the lack of recurrence of blurry LR images, ZSSR is less effective for SR, when facing very blurry LR images. By taking into this issue, Zhang \textit{et al.} \cite{zhang2019deep} proposed a deep plug-and-play SR framework for LR images with arbitrary blur kernels. This modified framework is flexible and effective to deal with very blurry LR images. Recent trends in SR also included SR for stereo images \cite{wang2019learning} and 3D appearance \cite{li20193d}.

\begin{figure*}[htbp]
	\centering
	\subfloat[The super-resolution results;]
	{
		\begin{minipage}[t]{0.3\textwidth}
			\centering
			\includegraphics[width=1\textwidth, height = 2.5cm]{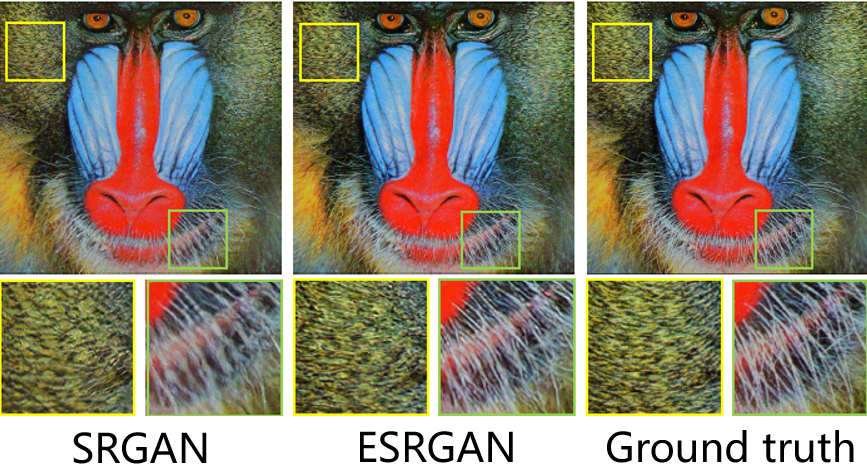}
			%\caption{Generative adversarial net;}
		\end{minipage}	
	}
	\subfloat[Image deblurring results;]
	{
		\begin{minipage}[t]{0.5\textwidth}
			\centering
			\includegraphics[width=1\textwidth, height = 2.5cm]{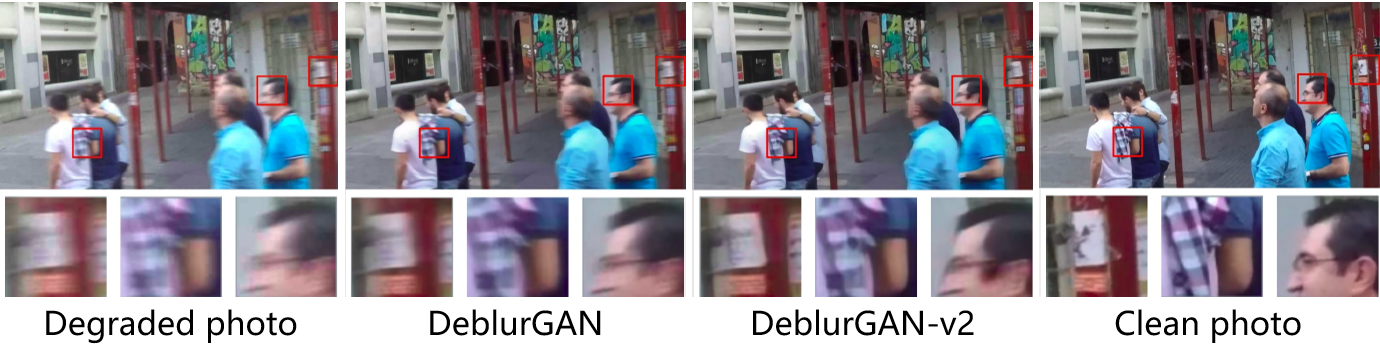}
			%\caption{Conditional generative adversarial net.}
		\end{minipage}
	}
	
	\subfloat[Image dehazing results;]
	{
		\begin{minipage}[t]{0.4\textwidth}
			\centering
			\includegraphics[width=1\textwidth, height = 2.5cm]{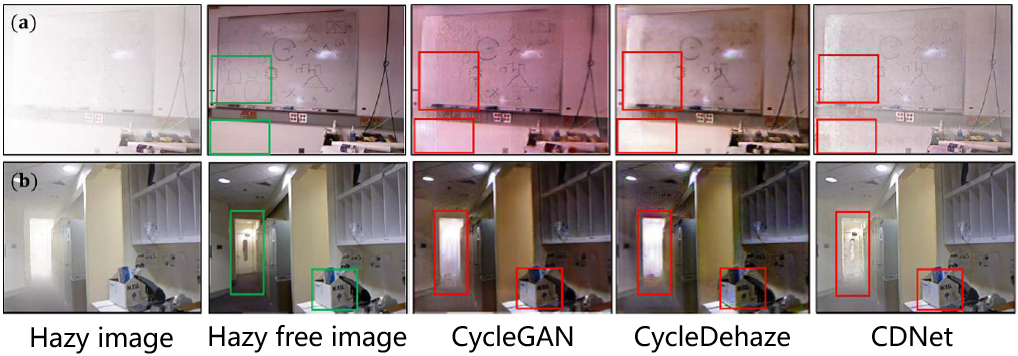}
			%\caption{Generative adversarial net;}
		\end{minipage}	
	}
	\subfloat[Image rain removal results.]
	{
		\begin{minipage}[t]{0.5\textwidth}
			\centering
			\includegraphics[width=1\textwidth, height = 2.5cm]{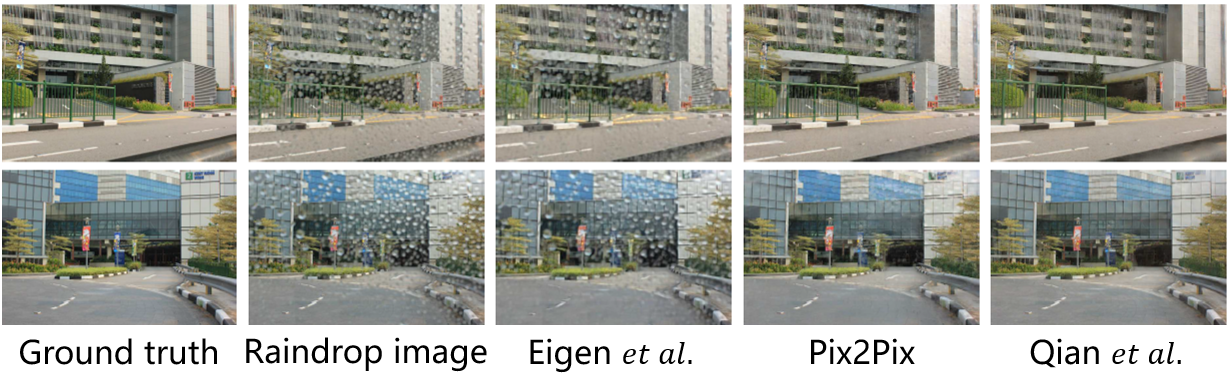}
			%\caption{Generative adversarial net;}
		\end{minipage}	
	}
	\caption{Generative adversarial networks for SR and Image deblurring/dehazing/rain removal. (a). The super-resolution results of $\times4$ for SRGAN, ESRGAN and the ground-truth; Reprinted by permission from \cite{wang2018esrgan}. Copyright. (b). Image deblurring results \copyright (2019) IEEE. Reprinted, with permission, from \cite{kupyn2019deblurgan}; (c). Image dehazing results \copyright (2019) IEEE. Reprinted, with permission, from \cite{dudhane2019cdnet}; (d). Image rain removal results \copyright (2018) IEEE. Reprinted, with permission, from \cite{qian2018attentive}. See also Fig. \ref{SR} and Table \ref{table3}.}
	\label{SR}
\end{figure*}

\textbf{GANs-based SR.} In addition to the traditional DL-based SR methods, GANs show their promising results in SR. The use of GANs for SR has the advantage of bringing the generated results closer to the natural image manifold, which may improve the accuracy of the result \cite{ledig2017photo}. The representative work on GANs-based SR (SRGAN) was presented by Ledig \textit{et al.} \cite{ledig2017photo}, which combines a content loss with an adversarial loss by training a GAN. This method is capable of reconstructing photo-realistic natural images for a $4\times$ upscaling factors. Although the SRGAN achieves good SR results, the local matching of texture statistics is not considered, which may restrict the improvement of the SR results to some extent. By considering this point, Sajjadi \textit{et al.} \cite{sajjadi2017enhancenet} focused on creating realistic textures to achieve SR. They proposed EnhanceNet, which combines adversarial training, perceptual loss, and a newly proposed texture transfer loss to achieve high-resolution results with realistic textures. In order to further improve the accuracy of SRGAN, Wang \textit{et al.} \cite{wang2018esrgan} extended SRGAN to ESRGAN by introducing residual-in-residual dense block and improving the discriminator and a perceptual loss. ESRGAN consistently has a better visual quality and natural texture than \cite{ledig2017photo}, as shown in Fig. \ref{SR}. 

HR images are conducive to improving the accuracy of perception tasks in autonomous systems. In autonomous systems, more complicated situations may be encountered, such as HR datasets are unavailable or the input LR images are noisy and blurry, which means that SR cannot be achieved with paired data. Inspired by the cycle consistency of CycleGAN, Yuan \textit{et al.} \cite{yuan2018unsupervised} tackled these issues with a cycle-in-cycle network (CinCGAN), which consists of two CycleGANs. The first CycleGAN maps LR images to the clean LR space, in which the proper denoising/deblurring processing is implemented on the original LR input. Then they stacked another well-trained deep model to up-sample the intermediate results to the desired size. Finally, they used adversarial learning to fine-tune the network in an end-to-end manner. The second CycleGAN contains the first one to achieve the purpose of mapping from the original LR to the HR. CinCGAN achieves comparable results to the supervised method \cite{dong2016accelerating}. Most SR methods trained on synthetic datasets are not effective in the real-world. SRGAN and EnhanceNet increase the perceptual quality by enhancing textures, which may produce fake details and unnatural artifacts. Soh \textit{et al.} \cite{soh2019natural} focused on the naturalness of the results to reconstruct realistic HR images. Further considering the transferability of the model, in order to solve the domain shift between synthetic data and real-world data, Gong \textit{et al.} \cite{gong2020learning} proposed to further minimize the domain gap by aligning the feature distribution while achieving SR. Specifically, they proposed a method to learn real-world SR images from a set of unpaired LR and HR images, which achieves satisfactory SR results on both paired and unpaired datasets. It is difficult to directly extend the image SR methods to the video SR. Recent developments included using the same framework to implement image SR and video SR \cite{brifman2019unified}, and real-time video SR using GANs \cite{lucas2019generative}.

Image super-resolution is used to increase the resolution of images, which helps to improve the accuracy of perception tasks. Although various SR models focus on improving accuracy, recent works have focused on the transferability of the model, like the transfer from synthetic datasets to real-world data. Future works may consider combining SR task with other tasks so that one model can achieve multi-task including SR. In addition to image SR task, we further consider image restoration, like image deblurring/dehazing/rain removal.

\subsection{Image deblurring \& Image dehazing \& Image rain removal}

Autonomous systems often encounter poor weather conditions, such as rain and fog, etc. There also exist blurry images due to poor shooting conditions or fast moving objects. It is well-recognized that the accuracy of computer vision tasks heavily depends on the quality of input images. Hence, it is of great importance to study image deblurring/dehazing/rain removal for autonomous systems, which make the high-level understanding tasks like semantic segmentation and depth estimation possible in practical applications of autonomous systems. It should be noted that although some image deblurring/dehazing/rain removal tasks use image enhancement algorithms \cite{joshi2010personal}, \cite{chen2015image}, \cite{fu2017clearing}, they are more relevant to image restoration than image enhancement. According to \cite{maini2010comprehensive}, image enhancement is to improve the viewer's perception of the image in a way that improves the information content, and it is designed to give emphasis to features of the image. Image restoration is to restore the noisy/corrupt image to its corresponding clean image, and the corruption may include motion blur and noise \cite{kuruvilla2016review}. Therefore, image deblurring/dehazing/rain removal can be regarded as image restoration tasks. When adversarial learning, like GANs, is used for image deblurring/dehazing/rain removal tasks, it can not only generate realistic images to improve the accuracy of image restoration, but also improve the transferability of the models, by considering the transfer from synthetic datasets to real-world images.

\textbf{Image deblurring.} Image blur is widely observed in autonomous systems, which heavily affects the understanding of the surroundings. In order to tackle the problem of image deblurring, several traditional DL-based methods without adversarial learning have been proposed successively \cite{sun2015learning}, \cite{schuler2015learning}, \cite{nah2017deep}. Considering the convincing performance of GANs in preserving image textures and creating realistic images, as well as inspired by image-to-image translation with GANs, Kupyn \textit{et al.} \cite{kupyn2018deblurgan} regarded image deblurring as a special image-to-image translation task. They proposed DeblurGAN, which is an end-to-end deblurring learning method based on cGANs. This method considers both accuracy and transferability, i.e., DeblurGAN improves deblurring results and it is 5 times faster than \cite{nah2017deep} for both synthetic and real-world blurry images. Then Kupyn \textit{et al.} \cite{kupyn2019deblurgan} futher improved DeblurGAN by adding a feature pyramid network to $G$ and adopting a double-scale $D$, which is called DeblurGAN-v2. DeblurGAN-v2 achieves better accuracy than DeblurGAN while being $10\sim100$ times faster than competitors, which will make it applicable to real-time video deblurring, as shown in Fig. \ref{SR}. Recently, Aljadaany \textit{et al.} \cite{aljadaany2019douglas} presented Dr-Net, which combines douglas-rachford iterations and Wasserstein-GAN \cite{gulrajani2017improved} to solve image deblurring without knowing the specific blurring kernel. In addition, Lu \textit{et al.} \cite{lu2019unsupervised} extracted the content and blur features separately from blurred images to encode the blur features accurately into the deblurring framework. They also utilized the cycle-consistency loss to preserve the content structure of the original images. Considering that stereo cameras are more commonly used in unmanned aerial vehicles, Zhou \textit{et al.} \cite{zhou2019davanet} focused the research on the deblurring of stereo images.

\textbf{Image dehazing.} Haze is a typical weather phenomenon with poor visibility, which forms a major obstacle for computer vision applications. Image dehazing is designed to recover clear scene reflections, atmospheric light colors, and transmission maps from input images \cite{engin2018cycle}. In recent years, a series of learning-based image dehazing methods have been proposed \cite{cai2016dehazenet}, \cite{ren2016single}, \cite{zhang2017joint}. Although these methods do not require prior information, their dependence on parameters and models may severely cause an impact on the quality of dehazing images. In order to reduce the effects of intermediate parameters on the model, and to establish an image dehazing method with good transferability, a series of GANs-based methods have been proposed for image dehazing. Li \textit{et al.} \cite{li2018single} tackled the image dehazing based on cGAN. Different from the basic cGAN, the generator in this method includes an encoder and decoder architecture, which helps the generator to capture more useful features to generate realistic results. The addition of cGAN makes the method in \cite{li2018single} achieve ideal results on both synthetic datasets and real-world hazy images. Considering the transferability of different scenarios and datasets, as well as independent of paired images, Engin \textit{et al.} \cite{engin2018cycle} proposed Cycle-Dehaze network by utilizing CycleGAN. This approach adds the cyclic perception-consistency loss and the cycle-consistency loss, thereby achieving image dehazing across datasets with unpaired images. Similar bidirectional GANs for dehazing have also been studied in \cite{kim2019bidirectional}. It is difficult for Cycle-Dehaze network to reconstruct real scene information without color distortion. Therefore, Dudhane \textit{et al.} \cite{dudhane2019cdnet} proposed the cycle-consistent generative adversarial network (CDNet), which utilized the optical model to find the haze distribution from the depth information. CDNet ensures that the fog-free scene is obtained without color distortion. The image dehazing results of Cycle-Dehaze and CDNet are shown in Fig. \ref{SR}. Most image dehazing methods only consider objects at the same scale-space, which will make dehazed images suffer from blurriness and halo artifacts. Sharma \textit{et al.} \cite{sharma2020scale} considered improving the accuracy and transferability of image dehazing, and presented an approach, which can remove haze based on per-pixel difference between Laplacians of Gaussian (LoG) of hazed images and original haze-free images at a scale-space. The model showed compelling results from simulated datasets to real-world maps, from indoor to outdoor. Recent developments in image dehazing also included targeting different channels, such as color channel \cite{ancuti2019color}, dark channel \cite{golts2019unsupervised}, and multi-scale networks \cite{liu2019griddehazenet}.

\textbf{Image rain removal.} Image rain removal is a challenging task, because the size, number and shape of raindrops are usually uncertain and difficult to learn. A number of methods have been proposed for image rain removal, but most of them require stereo image pairs \cite{yamashita2005removal}, image sequences \cite{yamashita2009noises}, or motion-based images \cite{you2015adherent}. Eigen \textit{et al.} \cite{eigen2013restoring} proposed a single image rain removal method, which is limited to dealing with relatively sparse and small raindrops. 

In order to improve the accuracy of the image rain removal results, consider the outstanding performance of GANs in the image inpainting or completion problems, a series of GANs-based methods have been used for image rain removal. Qian \textit{et al.} \cite{qian2018attentive} tackled the heavy raindrop removal from a single image using an attentive GAN. This method uses an attention map in both the generator and the discriminator. The generator produces an attention map through an attention-recurrent network and generates a raindrop-free image together with the input image. The discriminator evaluates the validity of the generation both globally and locally. The rain removal results of \cite{eigen2013restoring} and \cite{qian2018attentive} are shown in Fig. \ref{SR}. Nevertheless, this method is not suitable for torrential rain removal and is limited to raindrop removal. Considering heavy rain weather, strongly visible streaks or dense rain accumulations make the scene less visible. Li \textit{et al.} \cite{li2019heavy} considered the heavy rain situation and introduced an integrated two-stage CNN, which is able to remove rain streaks and rain accumulation simultaneously. In the first physics-based stage, a streak-aware decomposition module was proposed to decompose entangled rain streaks and rain accumulation to extract joint features. The second refinement stage utilized a cGAN that inputs the reconstructed map of the previous level and generates the final clean image. This method considered the transferability between the synthetic datasets and real-world images, and has achieved convincing results in both synthetic and real heavy rain scenarios. In order to improve the stability of GANs and reduce artifacts introduced by GANs in the output images, Zhang \textit{et al.} \cite{zhang2019image} proposed an image de-raining conditional generative adversarial network (ID-CGAN), which uses a multi-scale discriminator to leverage features from different scales to determine whether the de-rained image is from real data or generated ones. ID-CGAN has obtained satisfactory image rain removal results on both the synthetic dataset and real-world images. Jin \textit{et al.} \cite{jin2020ai} considered that existing methods may cause over-smoothing in derained images, and therefore they solved the problem from the perspective of feature disentanglement. They introduced an asynchronous interactive generative adversarial network (AI-GAN), which not only has achieved good results of image rain removal, but also has strong generalization capabilities, which can be used for image/video encoding, action recognition and person re-ID. 

Image deblurring/dehazing/rain removal tasks help to extract more useful information from bad weather scenes, which can help autonomous systems perceive the scene better. We focus on introducing GANs-based models, which improve the accuracy or transferability or both of them of these tasks. Future works may include the image deblurring/dehazing/rain removal tasks as the premise of perception, and then integrate a deeper model to achieve scene perception. After the introduction of tasks include image style transfer, image SR, and image deblurring/dehazing/rain removal, we consider high-level perception tasks like semantic segmentation.

\subsection{Semantic segmentation}

In emerging autonomous systems, such as autonomous driving and indoor navigation, scene understanding is required by means of semantic segmentation. Semantic segmentation is a pixel-level prediction method that can classify each pixel into different categories corresponding to their labels, such as airplanes, cars, traffic signs, or even backgrounds \cite{garcia2017review}. In addition, instance segmentation combines semantic segmentation and object detection to further distinguish object categories in the scene \cite{he2017mask}. Some traditional DL-based methods without adversarial learning have been proposed and have achieved good accuracy of semantic segmentation \cite{badrinarayanan2017segnet}, \cite{zhang2018fully} and instance segmentation \cite{he2017mask}, \cite{hu2018learning}. In practice, such annotations of pixel-level semantic information are usually expensive to obtain. Considering that the semantic labels of synthetic datasets are easy to obtain, it is helpful to consider semantic segmentation on labeled synthetic datasets and then transfer the results to real-world applications. Due to the domain shift between synthetic datasets and real-world images, it is worth exploring how to transfer the model trained on synthetic datasets to real-world images. By considering this point, adversarial learning is used to implement domain adaptation to improve the transferability of the model. Like other computer vision tasks in this review, the trend is now moving from improving accuracy to enhancing transferability. In this subsection, we focus on accuracy or transferability or both of them to review semantic segmentation and instance segmentation tasks.

\textbf{Traditional DL-based semantic segmentation.} Traditional DL-based semantic segmentation algorithms are mainly based on end-to-end convolutional network frameworks. To the best of our knowledge, Long \textit{et al.} \cite{long2015fully} were the first to train an end-to-end fully convolutional network (FCN) for semantic segmentation. The main insight is to replace fully connected layers with fully convolutional layers to output spatial maps. In addition, they defined a skip architecture to enhance the segmentation results. More importantly, the framework is suitable for input images of arbitrary size and can produce the correspondingly-sized output. This work is well-recognized as a milestone for semantic segmentation using DL. However, because the encoder network of this method has a large number of trainable parameters, the overall size of the network is large, which results in the difficulty to train FCN. Badrinarayanan \textit{et al.} \cite{badrinarayanan2017segnet} proposed SegNet, which has significantly fewer trainable parameters and can be trained in an end-to-end manner using SGD. SegNet is important in that the decoder performs the non-linear upsampling using the pooling index computed in the max-pooling step of the corresponding encoder, which eliminates the need to learn upsampling. Based on the encoder-decoder network of SegNet, DeepLab uses multi-scale contextual information to enrich semantic information. DeepLab proposed a series of semantic segmentation methods, like DeepLabv3+ \cite{chen2018encoder}, that combines a spatial pyramid pooling module and an encoder-decoder structure for semantic segmentation. In addition, the depthwise separable convolution is applied to both atrous spatial pyramid pooling and the decoder module to make the encoder-decoder network faster and stronger. 

\begin{figure*}[htbp]
	\centering
	\subfloat[CycleGAN semantic segmentation results;]
	{
		\begin{minipage}[t]{0.3\textwidth}
			\centering
			\includegraphics[width=5cm, height = 3cm]{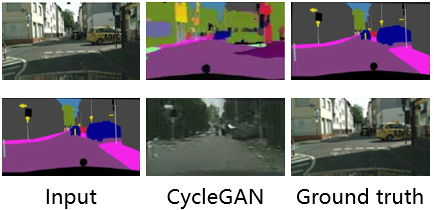}
			%\caption{Generative adversarial net;}
		\end{minipage}	
	}
	\subfloat[Experimental results on adaptation from cities in SYNTHIA fall to cities in SYNTHIA winter;]
	{
		\begin{minipage}[t]{0.35\textwidth}
			\centering
			\includegraphics[width=7cm, height = 3cm]{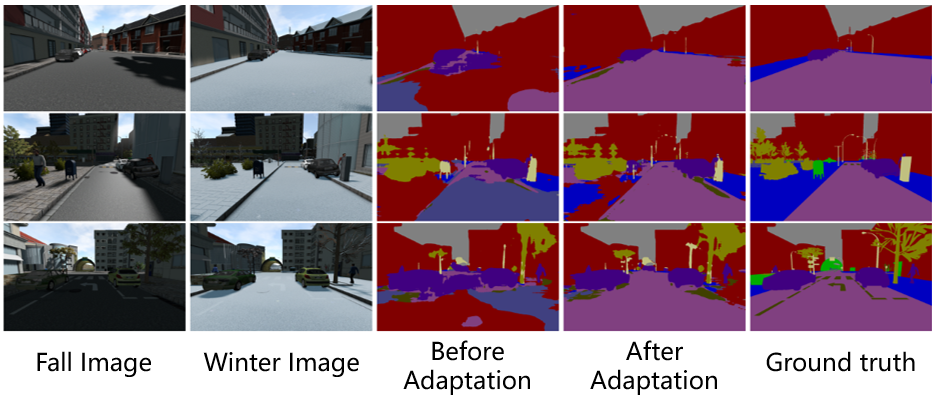}
			%\caption{Conditional generative adversarial net.}
		\end{minipage}
	}
	\subfloat[Multi-tasks.]
	{
		\begin{minipage}[t]{0.35\textwidth}
			\centering
			\includegraphics[width=4.3cm, height = 3cm]{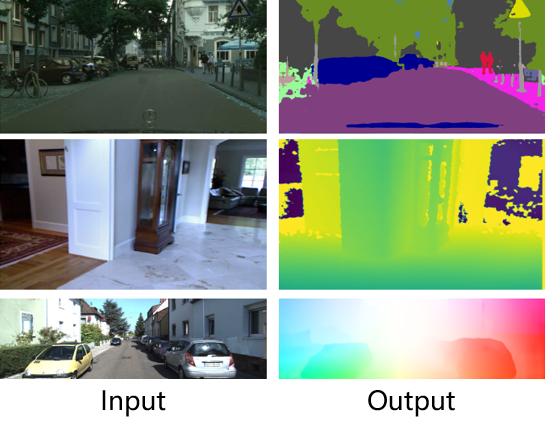}
			%\caption{Generative adversarial net;}
		\end{minipage}	
	}
	\caption{Generative adversarial networks for semantic segmentation and multi-task. (a). CycleGAN for semantic segmentation \copyright (2017) IEEE. Reprinted, with permission, from \cite{zhu2017unpaired}; (b). Qualitative results on adaptation from cities in SYNTHIA fall to cities in SYNTHIA winter \cite{hoffman2016fcns}; (c). Multi-task includes semantic segmentation (top row), depth prediction (middle row), to optical flow estimation (bottom row) \copyright (2019) IEEE. Reprinted, with permission, from \cite{chen2019crdoco}. See also Fig. \ref{semantic segmentation} and Table \ref{table3}.}
	\label{semantic segmentation}
\end{figure*}

The accuracy of unsupervised semantic segmentation tasks is usually worse than that of supervised methods, while supervised semantic segmentation often requires a lot of manual labeling, which is very costly. Note that a synthetic dataset with computer simulation like Grand Thief Auto (GTA) \cite{richter2016playing} can automatically label a large number of semantic tags, which is very important to improve the accuracy of the semantic segmentation model. However, due to the domain shift between the synthetic dataset and the real-world scene, it is necessary to consider domain adaptation in the semantic segmentation task. In order to address the domain gap problem and improve the transferability of the model, Hoffman \textit{et al.} \cite{hoffman2016fcns} proposed a domain adaptation framework with FCN for semantic segmentation, as shown in Fig. \ref{semantic segmentation}. This method considers aligning both global and local features through some specific adaptation techniques. This method makes full use of the label information of the synthesized dataset, and successfully transfers the results from a synthetic dataset to the real scene, in which a satisfactory semantic segmentation result is achieved in practical applications. The same combination of FCN with domain adaptation for semantic segmentation was also presented in \cite{zhang2018fully}. Zhang \textit{et al.} \cite{zhang2018fully} presented fully convolutional adaptation networks (FCANs), which also successfully explored domain adaptation for semantic segmentation. The model combines appearance adaptation networks and representation adaptation networks to synthesize images for domain adaptation at both the visual appearance-level and the representation-level. Recent developments also involved 3D semantic segmentation \cite{pham2019real}, \cite{liang2019hierarchical} and 3D instance segmentation \cite{lahoud20193d}.

\textbf{Traditional DL-based instance segmentation.} The more challenging task is \textit{instance segmentation}, which combines both object detection and semantic segmentation \cite{he2017mask}. Li \textit{et al.} \cite{li2017fully} first proposed an end-to-end fully convolutional method for instance-aware semantic segmentation. However, the method produced spurious edges on overlapping instances. He \textit{et al.} \cite{he2017mask} proposed Mask R-CNN, which is a classic instance segmentation algorithm. Mask R-CNN is easy to train and to generalize to other tasks, i.e., it performs breakthrough results in instance segmentation, bounding-box object detection, and person keypoint detection. This method includes two stages. The first stage proposes a candidate object bounding box. In the second stage, the prediction class and the box offset are in parallel, and the network outputs a binary mask for each region of interest (RoI). Mask R-CNN implements instance segmentation in a supervised manner, which is very expensive to semantic labels. In view of this, Hu \textit{et al.} \cite{hu2018learning} proposed a solution to a large-scale instance segmentation by developing a partially supervised learning paradigm, in which only a small part of the training process has instance masks, and the rest have box annotations. This method has demonstrated exciting new research directions in large-scale instance segmentation.

\textbf{GANs-based semantic segmentation.} GANs are flexible enough to reduce the differences between the segmentation result and the ground truth, and further improve the accuracy of the semantic segmentation results without manual labeling in some cases \cite{luc2016semantic}. As for using GANs for semantic segmentation, the typical methods are Pix2Pix \cite{isola2017image} and CycleGAN \cite{zhu2017unpaired}. The semantic segmentation result for CycleGAN is shown in Fig. \ref{semantic segmentation}. There are several variants based on Pix2Pix and CycleGAN, such as \cite{hoffman2017cycada},  \cite{liu2017unsupervised}, \cite{murez2018image}, etc. These methods do not only achieve satisfactory results in image style transfer, but also work well in semantic segmentation. Most of the adversarial domain adaptive semantic segmentation methods for subsequent improvements of CycleGAN and Pix2Pix improve the training stability and transferability by improving loss functions or network layers. Hong \textit{et al.} \cite{hong2018conditional} proposed a method based on cGAN for semantic segmentation. The network integrated cGAN into the FCN framework to reduce the gap between source and target domains. In practical tasks, objects often appear in an occluded state, which brings great challenges to the perception tasks of autonomous systems. To solve this problem, Ehsani \textit{et al.} \cite{ehsani2018segan} proposed SeGAN, that jointly generated the appearance and segmentation mask for invisible and visible regions of objects. Different from global alignment strategies like CycleGAN, Luo \textit{et al.} \cite{luo2019taking} further considered a joint distribution at the category-level. They proposed a category-level adversarial network (CLAN) to enhance local semantic consistency in the case of global feature alignment. Note that traditional semantic segmentation methods may suffer from the unsatisfactory quality of image-to-image conversion. Once the image-to-image conversion fails, nothing can be done to obtain satisfactory results in the subsequent stage of semantic segmentation. Li \textit{et al.} \cite{li2019bidirectional} tackled this problem by introducing a bidirectional learning framework with self-supervised learning, in which both translation and segmentation adaption models can promote each other in a closed loop. This segmentation adaptation model was trained on both synthetic and real-world datasets, which improved the segmentation performance of real-world data. In addition, Erkent \textit{et al.} \cite{erkent2020semantic} considered a method of semantic segmentation adapted to different weather conditions, which can achieve a satisfactory accuracy for semantic segmentation without the need of labeling the weather conditions of the source or target domain.

Semantic segmentation, as a high-level perception task of autonomous systems, predicts the semantic informantion of each pixel with a specific class label. Early supervised algorithms are expensive to collect labeled datasets from the real-world, and hence many algorithms consider the transferability between synthetic datasets and real-world data. Recent developments include semantic segmentation in more complex environments based on GANs, like bad weather conditions. Meanwhile, we think that instance segmentation based on GANs is also an open question. In addition to semantic segmentation, depth estimation is another high-level perception task of autonomous systems, which is very challenging to estimate the depth value of each pixel in the image.

\subsection{Depth estimation}

Depth estimation is an important task to help autonomous systems understand the 3D geometry of environments at high-level. A series of classical and learning-based methods were proposed to estimate depth based on motion \cite{ullman1979interpretation} or stereo images \cite{scharstein2002taxonomy}, which is computationally expensive. As widely known, due to the lack of complete scene 3D information, estimating the depth from a single image is an ill-posed task \cite{liu2015learning}. For the monocular depth estimation task, a series of traditional DL-based algorithms without adversarial learning have been proposed to improve the accuracy of the model. However, considering that it is expensive to collect well-annotated datasets in depth estimation tasks, it is appealing to use adversarial learning methods, like GANs, to achieve domain adaptation from synthetic datasets to real-world images. In addition, the adaptive method is used to improve the transferability of the model, so that the model trained on the synthetic dataset can be well transferred to real-world images. Here, we will introduce traditional DL-based depth estimation frameworks, as well as introduce methods to improve the transferability of depth estimation models by introducing adversarial learning.

%The traditional methods for depth estimation from a single image often require a lot of prior information \cite{ladicky2014pulling}.

\textbf{Traditional DL-based depth estimation.} Traditional DL-based depth estimation methods mainly focus on improving the accuracy of the results by using deep convolution frameworks. Eigen \textit{et al.} \cite{eigen2014depth} first proposed using a neural network to estimate depth from a single image in an end-to-end manner, which pioneeringly showed that it is promising for neural networks to estimate the depth from a single image. This framework consists of two components: the first one roughly estimated the global depth structure, and the second one refined this global prediction using local information. Considering the continuous property of the monocular depth value, depth estimation is transformed into a learning problem of a continuous conditional random field (CRF). Liu \textit{et al.} \cite{liu2015learning} presented a deep convolutional neural field model for single monocular depth estimation, which combined deep CNN and continuous CRF. This method achieved good results on both indoor and outdoor datasets. In order to reduce the dependence on the supervised signal and improve the transferability between different domains, unsupervised domain adaptation methods were presented for depth estimation in \cite{nath2018adadepth}. Some other developments in considering optical flow, camera pose and intrinsic parameters from monocular video for depth estimation can be found in \cite{chen2019self}. By considering the intrinsic parameters of the camera similar to \cite{gordon2019depth}, accurate depth information can be extracted from any video. 

\textbf{GANs-based depth estimation.} For the depth estimation task, it is too expensive to collect well-annotated image datasets. An appealing alternative is to use the unsupervised domain adaptation method via GANs to achieve domain adaptation from synthetic datasets to real-world images. Atapour-Abarghouei \textit{et al.} \cite{atapour2018real} took advantage of the adversarial domain adaptation to train a depth estimation model in a synthetic city environment and transferred it to the real scene. The framework consists of two stages. At the first stage, a depth estimation model is trained with the dataset captured in the virtual environment. At the second stage, the proposed method transfers synthetic style images into real-world ones to reduce the domain discrepancy. Although this method considers the transfer of synthetic city environment to the real-world scene, it ignores the specific geometric structure of the image in the target domain, which is important for improving the accuracy of depth estimation. Motivated by this problem, Zhao \textit{et al.} \cite{zhao2019geometry} proposed a geometry-aware symmetric domain adaptation network (GASDA), which produces high-quality results for both image style transfer and depth estimation. GASDA is based on CycleGAN \cite{zhu2017unpaired}, which performs translations for both synthetic-realistic and realistic-synthetic simultaneously with a geometric consistency loss of real stereo images. Zhao \textit{et al.} \cite{zhao2020domain} further considered high-level domain transformation, that is, mixing a large number of synthetic images with a small amount of real-world images. They proposed the attend-remove-complete (ARC) method, which learns to attend, remove and complete some challenging regions. The ARC method can ultimately make good use of synthetic data to generate accurate depth estimates. 

\textbf{Depth estimation via joint tasks learning.} Each pixel in one image usually contains surface normal orientation vector information and semantic labels, and both surface normal prediction, semantic segmentation and depth estimation are related to the geometry of objects, which makes it possible to train different structured prediction tasks in a consistent manner. To the best of our knowledge, there are some works that apply a single model to multiple related tasks. Note that for different tasks, the model should be fine-tuned \cite{eigen2015predicting}, \cite{mousavian2016joint} or the usage of different loss functions \cite{chen2019crdoco}, \cite{hwang2019adversarial}. Applying a single model to multiple related tasks through fine-tuning or using different loss functions shows that the model has a good transferability. Eigen \textit{et al.} \cite{eigen2015predicting} developed a more general network for depth estimation and applied it to other computer vision tasks, such as surface normal estimation and per-pixel semantic labeling. It is worth noting that Eigen \textit{et al.} used a single framework for depth estimation, surface normal estimation and semantic segmentation with only fine-tuning, which improved the framework of \cite{eigen2014depth} by considering a third scale at a higher resolution. Considering that GANs perform well in structured prediction space, Hwang \textit{et al.} \cite{hwang2019adversarial} proposed adversarial structure matching (ASM), which trains a structured prediction network through an adversarial process. This method achieved ideal results on monocular depth estimation, semantic segmentation and surface normal prediction. Although the ASM model has a good transferability for multiple tasks through different loss functions, its limitation is that specified datasets should be used for specific tasks, and it cannot be generalized to other datasets. To solve this limitation, Chen \textit{et al.} \cite{chen2019crdoco} embedded the pixel-level domain adaptation into the depth estimation task. Specifically, they proposed CrDoCo, a pixel-level adversarial domain adaptive algorithm for dense prediction tasks. The core idea of this method is that although the image styles of two domains may be different during the domain transfer process, the task predictions (e.g., depth estimation) should be exactly the same. Since CrDoCo is a pixel-level framework for dense prediction, it can be applied to semantic segmentation, depth prediction, and optical flow estimation, as shown in Fig. \ref{semantic segmentation}. CrDoCo can be applied to multi-task only by adjusting its loss function, and it also shows a good transferability between different datasets for a specific task.

Depth estimation helps autonomous systems understand the 3D structure of the surrounding scene. The transferability of depth estimation includes not only the transfer of synthetic to real-world data, but also the transfer of indoor to outdoor environments. Since depth, surface normals, and semantic labels are all related to object geometric information, recent works have also considered improving the accuracy of depth estimation by utilizing the interconnection between different tasks. We believe that future works should include the consideration of depth estimation under poor weather and light conditions. After reviewing the above-mentioned autonomous systems perception tasks, we will introduce pedestrian detection, re-identification and tracking tasks involved in autonomous systems.

\subsection{Pedestrian detection, re-identification and tracking}

Pedestrian detection, re-identification (re-ID) and tracking are very important for autonomous systems, especially for autonomous driving. The related works of pedestrian detection mainly focused on improving the accuracy of the results. Various developments have been made on improving the accuracy of complex visual environments, such as nighttime and occlusion. Person re-ID, which is more complicated than pedestrian detection, requires matching pedestrians in disjoint camera views. Traditional learning-based methods of person re-ID mainly focused on improving the accuracy of results, while recent GANs-based algorithms focused on transferability between domains. To further complicate the person re-ID, some developments consider locating targets in a sequence of time, that is, video tracking. The RL-based pedestrian tracking methods focus on not only accuracy but also transferability of the algorithms. In this subsection, we review pedestrian detection, re-ID and tracking tasks, focusing on accuracy or transferability or both of them.

\textbf{Pedestrian detection.} In recent years, pedestrian detection has been widely taken into account in autonomous systems, especially for autonomous driving and robot movement \cite{leibe2005pedestrian}, \cite{tian2015deep}. Pedestrian detection methods are generally divided into two categories: models based on hand-crafted features and deep models \cite{tian2015pedestrian}. Various models based on hand-crafted features have been proposed in the past few decades \cite{dalal2005histograms}, \cite{nam2014local}, \cite{cao2016pedestrian}. Although these models have made good progress, models based on hand-crafted features fail to extract semantic information. Sermanet \textit{et al.} \cite{sermanet2013pedestrian} used sparse convolutional feature hierarchies for pedestrian detection, which is named as ConvNet. The network first performs layer-wise training on the whole multi-stage system, and then uses the labeled data to fine-tune the complete architecture for the detection task. Although ConvNet learns features from training data, it treats pedestrian detection as a single binary classification task, which may confuse positive and negative samples. Therefore, Tian \textit{et al.} \cite{tian2015pedestrian} proposed a task-assistant CNN (TA-CNN), which can learn features from multiple tasks and multiple datasets. TA-CNN combines semantic tasks, including pedestrian attributes and scene attributes, to optimize pedestrian detection results. In order to further improve the accuracy of pedestrian detection in natural scenes, Li \textit{et al.} \cite{li2017scale} considered that the problem of large variance in pedestrian scale with different spatial scales may cause dramatically different features. Therefore, Li \textit{et al.} \cite{li2017scale} developed a scale-aware fast R-CNN (SAF R-CNN) framework, which combines a large-size sub-network and a small-size sub-network, as well as using the scale-aware weighting mechanism to deal with various sizes pedestrian in scenes. Although SAF R-CNN can detect pedestrian instances of different scales, it does not consider the factors like illumination conditions. In order to solve the problem of pedestrian detection in challenging illumination conditions at nighttime, Kim \textit{et al.} \cite{kim2019unpaired} used adversarial learning for cross-spectral pedestrian detection with unpaired setting. This method makes the color and thermal features of prominent areas where pedestrians exists to be similar by using adversarial learning, thereby improving the accuracy of pedestrian detection results at nighttime. Recent developments in pedestrian detection include tiny-scale pedestrian detection \cite{yin2019multi} and occluded pedestrian detection \cite{xie2020psc}, etc.

%Ouyang \textit{et al.} \cite{ouyang2013joint} proposed a model to jointly learn feature extraction, deformation handling, occlusion handling and classification, which are four important components in pedestrian detection. The interaction among the four components in this model allows them to maximize their strengths, which reduces the average miss rate of the experimental results. 

\textbf{Person re-identification.} A similar while more difficult task than pedestrian detection, person re-identification (re-ID), requires matching pedestrians in disjoint camera views. At present, there are several learning-based methods focusing on person re-ID \cite{ye2017dynamic}, \cite{wang2018transferable}, \cite{fan2018unsupervised}. However, these methods have poor transferability, that is, the person re-ID models trained on one domain usually fail to generalize well to another domain. Considering that CycleGAN shows great results in transferability using unpaired images, Deng \textit{et al.} \cite{deng2018image} introduced the similarity preserving cycle-consistent generative adversarial network (SPGAN), an unsupervised domain adaptation approach to generate samples while do not only have the target domain style but also preserve the underlying ID information. This method showed that applying domain adaptation to person re-ID can achieve competitive accuracy. Taking into account the data augmentation of different cameras, Zhong \textit{et al.} \cite{zhong2018camera} introduced the camera style (CamStyle) adaptation. CamStyle smooths disparities in camera styles, transferring labeled training image styles to each camera to augment the training set. CamStyle helps to learn pedestrian descriptors through camera-invariant property to improve re-ID experimental accuracy. The above approaches, like SPGAN \cite{deng2018image} and CamStyle \cite{zhong2018camera}, treated the domain gap as a black box and attempted to solve it by using a single style transformer. Liu \textit{et al.} \cite{liu2019adaptive} proposed a novel adaptive transfer network (ATNet), which investigates the root causes of the domain gap. ATNet realizes the domain transfer of person re-ID by decomposing complicated cross-domain transfers and transferring features through sub-GANs separately. Recently, Song \textit{et al.} \cite{song2020unsupervised} theoretically analyzed unsupervised domain adaptation re-ID tasks, which bridges the gap between theories of unsupervised domain adaptation and re-ID task. Recent developments in person re-ID also involved considering occluded parts \cite{hou2019vrstc} and different visual factors such as viewpoint, pose, illumination, and background \cite{sun2019dissecting}.

\begin{figure*}[htbp]
	\centering
%	\subfloat[Results of day$\rightarrow$night translation;]
	{
		\begin{minipage}[t]{0.9\textwidth}
			\centering
			\includegraphics[width=1\textwidth, height = 4.5cm]{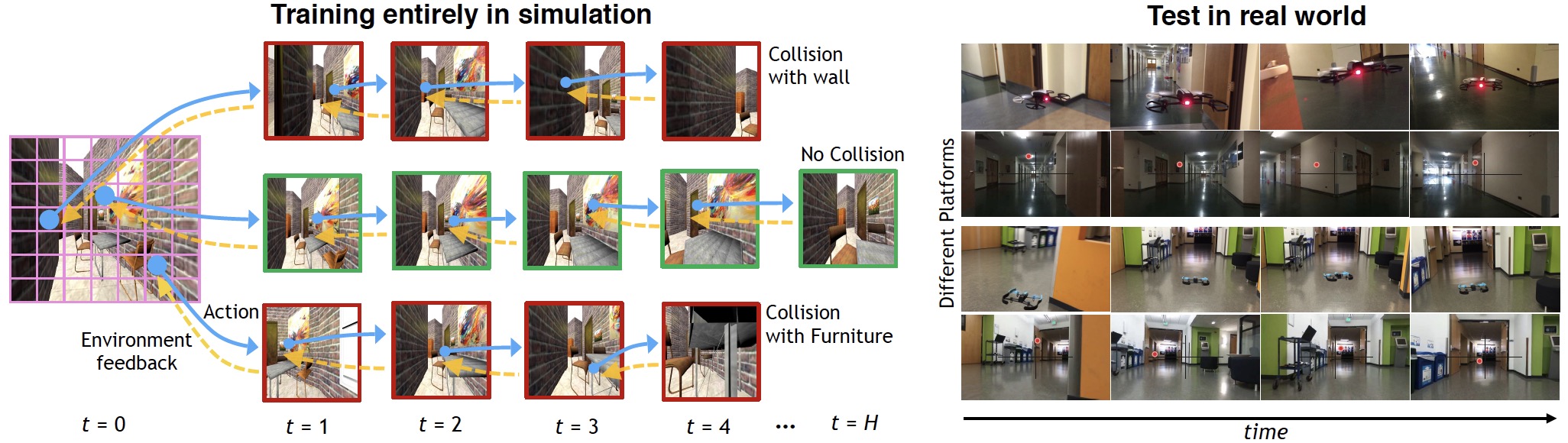}
		\end{minipage}	
	}\hspace{5mm}

	\caption{UAV navigation via DRL algorithm for indoor flying, which is entirely trained in a simulated 3D CAD model and generalized to real indoor flight environment \cite{sadeghi2016cad2rl}. See also Fig. \ref{UAV navigation} and Table \ref{table4}.}
	\label{UAV navigation}
\end{figure*}

\textbf{Pedestrian tracking.} 
Video tracking is an improvement of person re-ID and needs to locate the target in a sequence of time, which is difficult to handle because of tracking obstacles. When it comes to accuracy and transferability, RL-based methods in pedestrian tracking are concerned with whether the action in each frame is discrete or continuous and whether the labelled bounding-boxes at each frame is limited or not. Tracking pedestrian by searching a series of discrete actions in each frame is a solution. Yun \textit{et al.} \cite{yun2017action} proposed the action-decision network (ADNet) to generate actions to find the location and the size of the target object in a new frame. The ADNet is updated by performing tracking simulation on the training sequence and utilizing action dynamics with the help of RL. After pre-training ADNet by superivsed learning, online adaptation is applied to the network to accommodate the appearance changes or deformation of the target during tracking test sequences. Therefore, the pre-trained ADNet features can be transferred to a new frame during online adaption.
When the bounding-boxes at each frame is limited, the algorithm can also be trained successfully and transferred to new frames with the help of larger training datasets. Supancic \textit{et al.} \cite{supancic2017tracking} used RL to train trackers with more limited supervision on far more massive datasets. The results illustrated that the algorithm can track pedestrian on a never-before-seen video, the video can be used for both evaluation of the current tracker and for training the tracker for future use. In a word, the learning structure is informative and the features contained in video will be transferred to another training process.
Furthermore, in order to exploit continuous actions for visual tracing, which improves training efficiency and accuracy, Chen \textit{et al.}  \cite{chen2018real} put forward a real time AC framework to exploit continuous action space for visual tracking. For online tracking, the ‘actor’ model provides an offline dynamic search strategy to locate the target object in each frame efficiently by only one action output, and the ‘critic’ model acts as a verification module to make the tracker more robust. The real-time performance of the trackers is better than state-of-the-art methods, like MDNet \cite{nam2016learning} and ADNet \cite{yun2017action}. Similar to \cite{chen2018real}, Luo \textit{et al.} \cite{luo2019end} used the same RL method to deal with continuous tracking problems. Moreover, they introduced an environment augmentation technique, that is, virtual environments named ViZDoom \cite{kempka2016vizdoom}, in order to boost the tracker’s generalization ability.

The current methods of pedestrian detection focus on improving the accuracy of the detection results, while the GANs-based person re-ID methods mentioned in this survey center on improving the transferability of the algorithm. The RL-based pedestrian tracking methods concentrate on equipping both the accuracy and transferability. The future works should include pedestrian detection, re-ID and tracking in severe occlusion situations. In addition, future works may also involve pedestrian detection, re-ID and tracking tasks at the semantic level. In addition to perception tasks, we will introduce some decision-making tasks, like robot navigation. Robot navigation focuses on navigating robot to avoid collision or to a target considering accuracy or transferability or both of them of tasks.

\subsection{Robot Navigation} \label{H}

Recently, robot navigation is a key and hot topic in autonomous systems, which mainly focuses on navigating robot to a target position or to avoid obstacles in a known/unknown environment. In this review, we consider whether the trained model can accurately learn the task feature or successfully transfer the previous information to new tasks or domains.
%As mentioned before, RL is capable of dealing with decision-making issues in a known/unknown environment. 
%Therefore, various studies use RL or meta-learning methods to learn policies in static and dynamic environment from observations and state information \cite{long2018towards}, \cite{everett2018motion}, 
A variety of RL and meta-learning methods, such as DQN \cite{zhang2017deep}, LSTM structure \cite{alahi2016social} and MAML \cite{gupta2018meta}, etc, can accurately or transferably handle the changes arising from the environment or task when using the previously trained model. As shown in Table \ref{table4}, we summarize the RL and meta-learning methods to handle accurate learning and domain-transfer tasks in robot and UAV navigation issues. As for experiment platform, AI2-THOR (the house of interactions) \cite{zhu2017target} performs well due to its shared task features and datasets, assuring the learned skills transfer to new tasks.
%RL methods tend to focus on transferring tasks from simulation to real world, while 
Moreover, meta-learning methods usually have more satisfactory transferability than RL methods when lacking training and testing data by means of extracting or memorizing previous training data in simulation.
%that will affect the training accuracy when using the previously trained model. Moreover, when dealing with new issues, the model

\begin{table*}
\scriptsize
\centering
	\caption{Summary of traditional RL/meta-learning methods for scenario-transfer tasks. (We classify the meta-learning methods into several classes. ``A" represents recurrent network. ``B" represents metric network. ``C" represents MAML. ``D" represents meta-imitation learning. ``E" represents meta-RL. Similarly, we classify the RL methods into several classes. ``F" represents Fitted Q-iteration. ``G" represents soft Q-learning. ``H" represents DQN. ``I" represents DDPG. ``J" represents soft AC. ``K" represents A3C. ``L" represents GPS. ``M" represents asynchronous NAF (normalized advantage function) \cite{gu2016continuous}. ``N" represents PPO (proximal policy optimization) \cite{schulman2017proximal}. ``O" represents TRPO (trust region policy optimization) \cite{schulman2015trust}. ``P" represents DPP.)}
	\begin{tabular}{cccccccc}
		\toprule
		Year & Reference & Task & RL method & Meta-learning method & Simulation platform & Practice platform \\
		\midrule
		2016 & Sadeghi \textit{et al.} \cite{sadeghi2016cad2rl} & UAV navigation & F &  & 3D CAD environment & Parrot Bebop \\ 
		2017 & Tai \textit{et al.} \cite{tai2017virtual} & Robot navigation & I &  & V-REP & Turtlebot   \\
		2017 & Zhang \textit{et al.} \cite{zhang2017deep} & Robot navigation & H &  & Maze-like 3D environment  & Robotino   \\
	%	2017 & Hwangbo \textit{et al.} \cite{hwangbo2017control} & UAV Navigation & RL & \tabincell{c}{ Deterministic \\ On-policy Method } & Robotic Artificial Intelligence & Humming Bird Quadrotor \\
		2017 & Polvara \textit{et al.} \cite{polvara2017autonomous} & UAV navigation & H &  & Gazebo & Parrot AR Drone 2   \\
		2017 & Zhu \textit{et al.} \cite{zhu2017target} & Robot navigation & K & B & AI2-THOR & SCITOS  \\
	%2018 & Kahn \textit{et al.} \cite{kahn2018self} & Robot navigation & Both & \tabincell{c}{RL with generalized \\ computation graphs and A} & Simulated RC car  & RC car   \\
		2018 & Banino \textit{et al.} \cite{banino2018vector} & Robot navigation & K & A & Multi-room 2D environment  & None   \\
		2018 & Faust \textit{et al.} \cite{faust2018prm} & Robot navigation & I &  & Simulated building plans  & Differential drive robot    \\
		2019 & Zhu \textit{et al.} \cite{zhu2019sim} & Robot navigation & K & A & SUNCG & Matterport3D   \\
		2019 & Niroui \textit{et al.} \cite{niroui2019deep} & Robot navigation & K & A & Turtlebot Stage simulator  & Turtlebot   \\
		2019 & Wortsman \textit{et al.} \cite{wortsman2019learning} & Robot navigation &  & A, C, E & AI2-THOR & None   \\
	    2019 & Jabri \textit{et al.} \cite{jabri2019unsupervised} & Robot navigation &  & E & ViZDoom & None   \\
	    2019 & Koch \textit{et al.} \cite{koch2019reinforcement} & UAV navigation & I, O, N & & GymFC & None   \\
		2020 & Gaudet \textit{et al.} \cite{gaudet2020adaptive} & UAV navigation &  & E & Mars and asteroid landing simualtion & None   \\
		\midrule
        2015 & Zhang \textit{et al.} \cite{zhang2015towards} & Robotic manipulation & H &  & None & Baxter arm \\
        2016 & Levine \textit{et al.} \cite{levine2016end} & Robotic manipulation & L &  & MuJoCo & PR2 robot   \\
        2017 & Gu \textit{et al.} \cite{gu2017deep} & Robotic manipulation & M &  & MuJoCo & 7-DoF arm \\
        2017 & Finn \textit{et al.} \cite{finn2017one} & Robotic manipulation &  &  C, D & MuJoCo & 7-DoF PR2 arm  \\
        2018 & Haarnoja \textit{et al.} \cite{haarnoja2018composable} & Robotic manipulation & G &  & MuJoCo & \tabincell{c}{ 7-DoF Sawyer arm} \\
		2018 & Zhu \textit{et al.} \cite{zhu2018reinforcement} & Robotic manipulation & N & A & MuJoCo & Jaco robot arm \\
		2018 & Zeng \textit{et al.} \cite{zeng2018learning} & Robotic manipulation & H &  & V-REP &  UR5 robot arm \textit{et al.} \\
	%	2018 & Peng \textit{et al.} \cite{peng2018deepmimic} & Action Imitation & PPO & Bullet pyhsics engine & None \\
	    	%	2018 & Peng \textit{et al.} \cite{peng2018deepmimic} & Action Imitation & PPO & Bullet pyhsics engine & None \\
        2018 & Yu \textit{et al.} \cite{yu2018one} & Robotic manipulation &  & C, D & MuJoCo & 7-DoF PR2 arm \textit{et al.} \\
        2019 & Yu \textit{et al.} \cite{yu2019meta} & Robotic manipulation & N, O, J & C & MuJoCo & None \\
        2019 & Zeng \textit{et al.} \cite{zeng2018robotic} & Robotic manipulation &  & B & None & Amazon Robotics Challenge \\
        2019 & Tsurumine \textit{et al.} \cite{tsurumine2019deep} & Robotic manipulation & P &  & N DOF simulated manipulator & 15-DoF humanoid robot \\
        2020 & Singh \textit{et al.} \cite{singh2020scalable} & Robotic manipulation &  & D & Bullet physics engine & None \\
		\bottomrule
	\end{tabular}
\label{table4}
\end{table*}

\textbf{RL-based robot navigation.}
In order to improve training efficiency and accuracy, dividing a single task to several sub-tasks and training them separately is a solution. Polvara \textit{et al.} \cite{polvara2017autonomous} proposed two distinct DQNs, called double DQNs, which were used to train two sub-tasks: landmark detection and vertical landing, respectively. Due to the separate training of each single task at the same time, training efficiency and accuracy were improved to an extent. Moreover, training the model with various \emph{auxiliary tasks}, such as pixel control \cite{bellemare2019geometric}, reward prediction \cite{jaderberg2016reinforcement} and value function replay \cite{mirowski2016learning}, will also help the robot adapting to the target faster and more accurate. 
%In \cite{mirowski2016learning}, Mirowski \textit{et al.} considered two auxiliary tasks in visual navigation issue. In the first task, a low-dimensional depth map is reconstructed at each time step, which is advantageous for obstacle avoidance and short-term path planning. The other task involves loopback detection, which is used to detect whether the current position has been accessed within the current trajectory. Experiments proved that the co-training with multi-tasks can significantly improve the training efficiency.

In order to equip the model with a better transferability when encountering a new situation, tasks features \cite{parisotto2015actor}, \cite{zhang2017deep} and training policies \cite{chen2018deep} can be transferred to novel tasks in the same domain or across domains. Parisotto \textit{et al.} \cite{parisotto2015actor} and Rusu \textit{et al.} \cite{rusu2015policy} transferred useful features among different ATARI games and then the corresponding features were utilized to train a new ATARI game in the same domain. In addition, when dealing with the tasks whose trials in real world are usually time-consuming or expensive, the characteristic of tasks can be transferred cross-domain effectively. Zhang \textit{et al.} \cite{zhang2017deep} put forward a shared DQN between tasks in order to learn informative features of tasks, which can be transferred from simulation to real world. Similarly, as shown in Fig. \ref{UAV navigation}, Sadeghi \textit{et al.} \cite{sadeghi2016cad2rl} proposed a novel realistic translation network, which transforms virtual image inputs into real images with a similar scene structure. Moreover, policies can be transferred from simulation to simulation. Similar to \cite{polvara2017autonomous}, the primary training policy of \cite{sadeghi2016cad2rl} can be divided into several secondary policies, which acquire certain behaviors. Then these behaviors are combined to train the primary policy, which helps to make the primary policy more transferable across domains. Chen \textit{et al.} \cite{chen2018deep} used AC networks to train the secondary policies as well as the primary policy. In navigation, the primary behavior learned by a high-degree-freedom robot is to navigate straightly to the target with a sample environment. Then Chen \textit{et al.} randomized the non-essential aspects of every secondary behavior, such as the appearance, the positions and the number of obstacles in the scene to improve generalization ability of the final policy. 

%Additionally, the abstracted input information will help to transfer information from simulation to real world. Tai \textit{et al.} \cite{tai2017virtual} took 10-dimensional sparse range findings as the observation input and train a mapless motion planer navigating a robot to a desired target and avoid collision. The abstracted input images are sampled from specific angles of the raw laser range results based on a trivial distribution, which reduces the gap in observation between the virtual and real environments. Therefore, the planer trained in simulation can be directly transferred to real world. 
Due to the sampling constraints of model-free RL methods and transferring limits of model-based RL methods as mentioned in Section II, it is difficult to equip a model with good transferability and sampling efficiency at the same time. An easy way to handle this contradiction is to combine model-free methods with model-based methods. Kahn \textit{et al.} \cite{kahn2018self} used a generalized computation graph to find the navigation policies from scratch by inserting specific instantiations between model-free and model-based ones. Therefore, the algorithm not only learns high-dimensional tasks but also has promising sampling efficiency. 
%Due to the limits of model-free and model-based methods as mentioned in Section II. Combining model-free methods with model-based methods is a feasible way to train a model with sufficient training data when dealing with complex tasks. Kahn \textit{et al.} \cite{kahn2018self} used a generalized computation graph to find the navigation policies from scratch by inserting specific instantiations between model-free and model-based methods. Therefore, the algorithm not only learns high-dimensional tasks but also has promising sampling efficiency.

\textbf{Meta-learning-based robot navigation.}
RL-based methods tend to need sufficient training data in order to acquire transferability. When a new task has insufficient data during training and testing, \emph{meta-learning} methods can also promote the model to be transferable across domains. Firstly, recurrent models, like LSTM structure, weaken the long-term dependency of sequential data, which acts as an optimizer to learn an optimization method for the gradient descent models. 
%In \cite{mirowski2016learning}, the authors combined the LSTM structure with auxiliary tasks and A3C networks to navigate in a maze to find a target using the images of the monocular camera. In real word navigation, training data are more various and unpredictable than those in simulation experiments. Therefore, the LSTM structure play a vital role in generating good navigation policies. 
Mirowski \textit{et al.} \cite{mirowski2018learning} proposed a multi-city navigation network with LSTM structure. The main task of the LSTM structure was used to encode and encapsulate region-specific features and structures in order to add multiple paths in each city. After training in multiple cities, it was proved that the network is sufficiently versatile. 
Moreover, metric learning can be utilized to extract image information and generalize the specific information, which is helpful in navigation. Zhu \textit{et al.} \cite{zhu2017target} combined siamese network with AC network to navigate the robot to the target only with 3D images. Siamese network captures and compares the special characteristics from the observation image and target image. Then, the joint representation of images is kept in scene-specific layers. AC network uses the features in scene-specific layers to generate policy and value outputs in navigation. To sum up, the deep siamese AC network shares parameters across different tasks and domains so that the model can be generalized across targets and scenes. 
Even if the models trained by the two meta-learning methods above acquire both accuracy and transferability, when the models encounter new cross-domain tasks, they also need plenty of data to be retrained. In order to fine-tune a new model with few data, MAML is a good way of thought. In \cite{finn2017model}, it was verified that MAML performs well in 2D navigation and locomotion simulation, compared with traditional policy gradient algorithms. It is shown that MAML could learn a model that adapts much more quickly in a single gradient update, while it continues to improve with additional updates without overfitting. When the training process is unsupervised, MAML is not applicable and needs to be adjusted, such as constructing a reward function during meta-training process \cite{gupta2018unsupervised} and labeling data using clustering methods \cite{hsu2018unsupervised}, etc. In \cite{wortsman2019learning}, Wortsman \textit{et al.} proposed a self-adaptive visual navigation (SAVN) method derived from MAML to learn adapting to new environments without any supervision. Specifically, SAVN optimizes two objective functions: self-supervised interaction loss and navigation loss. During training, the interaction- and navigation-gradients are back-propagated through the network, and the parameters of the self-supervised loss are updated at the end of each episode using navigation-gradients, which is trained by MAML. During testing, the parameters of the interaction loss remain fixed, while the rest of the network is updated using interaction-gradients. Therefore, the model equips the MAML methods with good transferability in a no supervision environment.

RL or meta-learning methods help the robot navigate to targets or avoid obstacles. When using RL methods, separating tasks or adding auxiliary tasks during training process will improve the accuracy of the navigation results. Moreover, there are many ways to improve transferability in robot navigation, including task-transfer, parameter-transfer and policy-transfer, etc. Compared with RL methods, meta-learning methods promote the transferability well especially when the training and testing data are limited. In the future, with the popularity of MAML, we believe that MAML is capable of handling more complex tasks in reality and achieving more satisfactory transferability by means of combining with some state-of-art methods, such as metric learning and LSTM structure, etc.
After introducing the above-mentioned autonomous systems robot navigation tasks, we will focus on another robotic issue, i.e., robotic manipulation.

\subsection{Robotic Manipulation}

%As mentioned above, model-free RL methods, such as Q-learning \cite{watkins1992q} and DQN \cite{mnih2013playing}, are able to handle more complex tasks with satisfactory transferability compared to model-based RL methods, such as GPS \cite{levine2016end} and MBVE \cite{feinberg2018model}. 
In this section, we will focus on transferability in robotic manipulation issues according to domain-transfer tasks implemented by various robotic arms. Compared with robot navigation, robotic manipulation mainly considers precise control of robotic arms by means of multiple degrees of freedom. RL methods enable the robotic arm to transfer between different environments and tasks by means of special inputs \cite{kompella2017continual} and reformed training networks \cite{andrychowicz2017hindsight}, etc.
Moreover, meta-learning and imitation learning can be utilized to handle difficult tasks with few or even one demonstration during meta testing process in the same domain or across domains \cite{finn2017one}, \cite{yu2018one}, in order to speed up the learning process and transfer previous task features. In Table \ref{table4}, we summarize RL and meta learning methods to deal with domain-transfer robotic manipulation problems. As for experiment platform, MuJoCo (multi-joint dynamics with contact) \cite{todorov2012mujoco} and PR2 arm are popular because robotic arms with multiple degrees of freedom and shared information have better accuracy and transferability. 
Moreover, compared with RL, meta-learning is capable of training with fewer data and adapting to new tasks faster to acquire the model transferability.

%MuJoCo (multi-joint dynamics with contact) \cite{todorov2012mujoco} simulation platform is popular 

%Therefore, RL and meta-learning are very helpful transferable methods in the field of robotic manipulation \cite{gu2017deep}.

%Combining deep neural network with Q-learning, DQN and many derivative networks from DQN are widely used in the field of robotic manipulation \cite{vecerik2017leveraging}, \cite{andrychowicz2017hindsight}.

\textbf{RL-based robotic manipulation.}
When considering improving the transferability of robotic arm systems, synthetic data as input \cite{zhang2015towards} and separate networks in training \cite{zhang2016modular} are possible RL-based solutions. Synthetic inputs help to transfer experience learned from different settings in simulation to real world. Zhang \textit{et al.} \cite{zhang2015towards} were the first to learn controlling a three-joint robot arm via DQN merely from raw-pixel images without any prior knowledge. The robot arm reaches the target in real world successfully, only when it takes synthetic images that generated by the 2D simulator as inputs according to real-time joint angles. Therefore, the input of synthetic images inevitably offsets the gap between the simulation and real world, thereby improving the transferability.
Moreover, when the data is limited and unable to be synthesized, DQN can be divided into perception and control modules, which are trained separately. Then, the perception skills and the controller obtained from simulation will be transferable \cite{zhang2016modular}. Similarly, DQN can also train several networks and combine the experience learned together. Zeng \textit{et al.} \cite{zeng2018learning} used DQN to jointly train two fully convolutional networks mapping from visual observations to actions. The experience will transfer between robot pushing and grasping processes, and thus these synergies are learned. To make a comparison of some popular RL methods focusing on generalization ability in robotic manipulation, Quillen \textit{et al.} \cite{quillen2018deep} evaluated simulated benchmark tasks, in which robot arms were used to grasp random targets in comparison with some DRL algorithms, such as double Q-learning (DQL), DDPG, path consistency learning (PCL), Monte Carlo (MC) policy evaluation. In the experiment, the trained robot arms coped with grasping unseen targets. The results revealed that DQL performs better than other algorithms in low-data regimes, and has a relatively higher robustness to the choice of hyperparameters. When data is becoming plentiful, MC policy evaluation achieves a slightly better performance.

\begin{figure*}[htbp]
	\centering
	\subfloat[Examples of reaching tasks;]
	{
		\begin{minipage}[t]{0.47\textwidth}
			\centering
			\includegraphics[width=0.98\textwidth, height = 4cm]{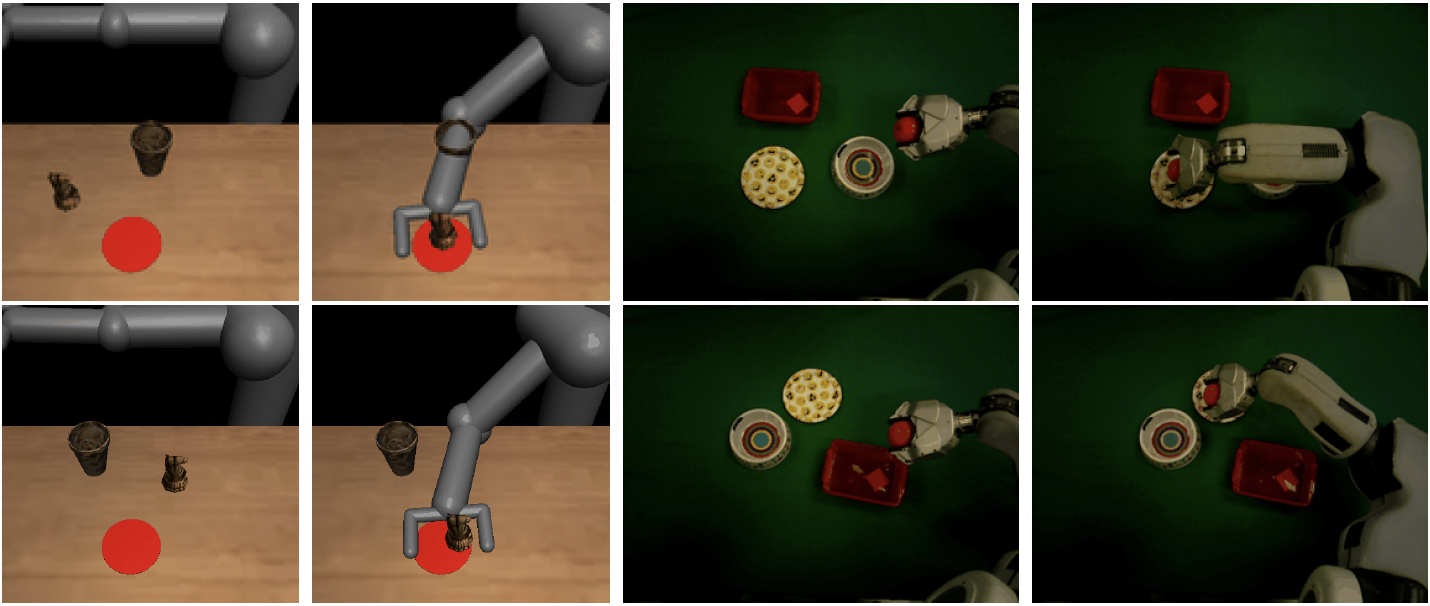}
			%\caption{Generative adversarial net;}
		\end{minipage}	
	}
	\subfloat[Examples of placing, pushing and pick-and-place tasks.]
	{
		\begin{minipage}[t]{0.47\textwidth}
			\centering
			\includegraphics[width=0.98\textwidth, height = 4cm]{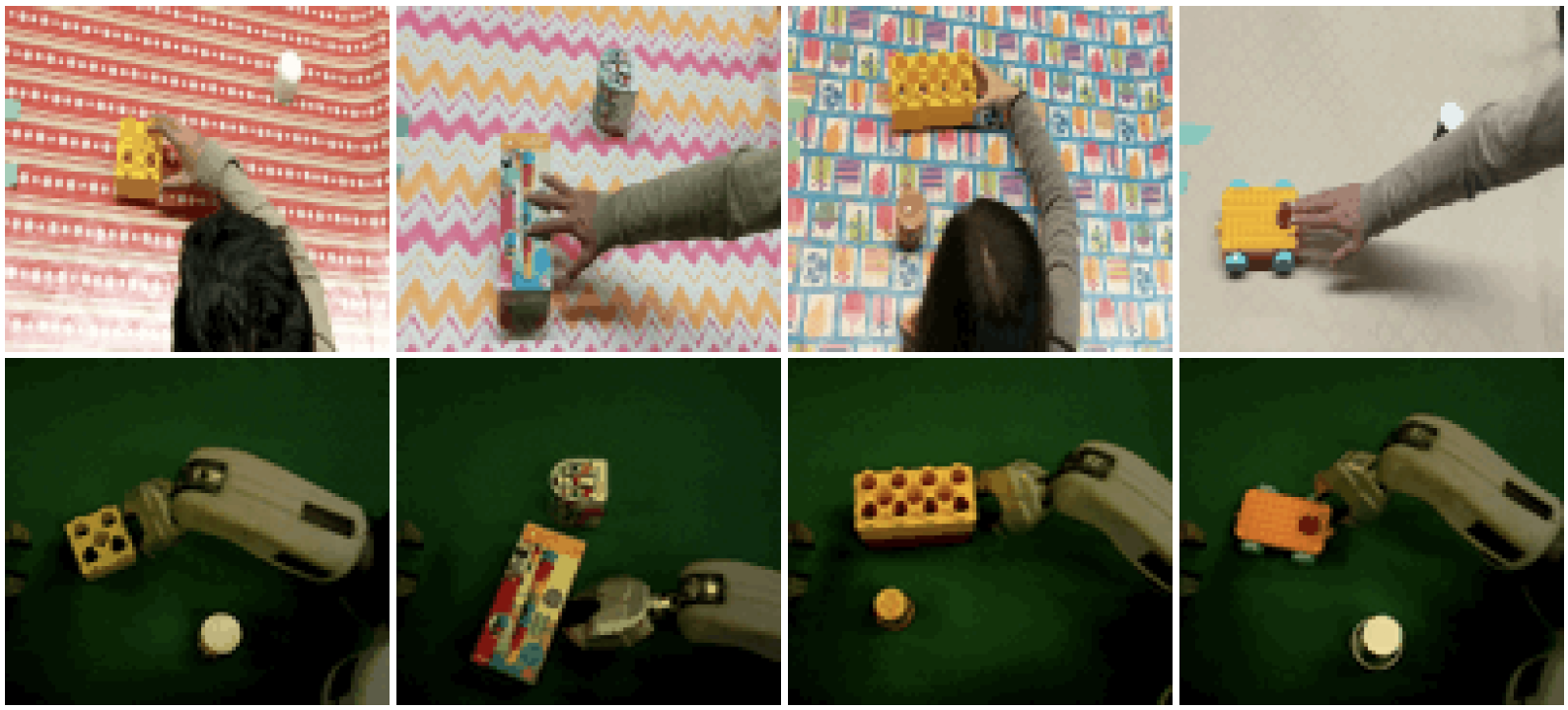}
			%\caption{Generative adversarial net;}
		\end{minipage}	
	}
	\caption{Demonstrations and robotic actions in simulation and real world. (a). Robot demonstrations used for meta-imitation learning \cite{finn2017one}; (b). Human and robot demonstrations used for meta-imitation learning with large domain shift \cite{yu2018one}. See also Fig. \ref{robotic actions} and Table \ref{table4}.}
	\label{robotic actions}
\end{figure*}

\textbf{MAML-based robotic manipulation.}
However, in robotic manipulation issues, traditional RL methods tend to need plenty of training data. Even if they can transfer to new tasks or domains, they also have poor generalization ability \cite{gu2017deep}, \cite{kalashnikov2018qt}. MAML combined with imitation learning, is able to utilize past experience across different tasks or domains, which can learn new skills from a very small number of demonstration in various fields of application. Duan \textit{et al.} \cite{duan2017one} let the robot arm demonstrate itself in simulation, that is, the input and output samples were collected by the robot arm itself. The inputs of the model are the position information of each block rather than images or videos. They first sampled a demonstration from one of the training tasks. Then, they sampled another pair of observation and action from a second demonstration of the same task. Considering both the first demonstration and second observation, the network was trained to output the corresponding action. In manipulation network, the soft attention structure allows the model to generalize to conditions and tasks that are invisible in training data. Then, Finn \textit{et al.} \cite{finn2017one} used visual inputs from raw pixels as demonstration. The model requires data from significantly fewer prior demonstrations in training and merely one demonstration in testing to learn new skills effectively. Moreover, it not only performs well in simulation but also works in real robotic arm system. MAML is modified to two-head architecture, which means that the algorithm is flexible for both learning to adapt policy parameters and learning the expert demonstration. Therefore, the number of demonstrations needed for an individual task is reduced by sharing the data across tasks. Taking robot arm pushing as an example, during the training process, the robot arm can see various pushing demonstrations, which contain different objects and each object may have different quality and friction, etc. In the testing process, the robot arm needs to push the object that has never seen during training. It needs to learn which object to push and how to push it according to merely one demonstration. As shown in Fig. \ref{robotic actions}, compared with \cite{finn2017one}, Yu \textit{et al.} \cite{yu2018one} increased the difficulty of imitation learning, that is, only using a single video demonstration from a human as input and the robot arm needs to accomplish the same work as \cite{finn2017one} by domain-adaption. The authors put forward a domain-adaptive meta-learning method that transfers the data from human demonstrations to robot arm demonstrations. MAML was utilized to deal with the setting of learning from video demonstrations of humans. Due to the clone of behavior across the domain, the loss function also needs to be reconstructed and TCN is used to
construct the loss network in MAML structure in the robotic arm domain. Specifically, the robot arm will learn a set of initial parameters in the video domain, then after one or a few steps of gradient descent on merely one human demonstration, the robot arm is able to perform the new task effectively. Recently, on the basis of \cite{finn2017one}, \cite{singh2020scalable} improved the one-shot imitation model by using additional autonomously-collected data instead of manually collecting data. It is novel that they put forward an embedding network to distinguish whether two demo embeddings are close to each other. By the use of metric learning, they compute the Euclidean distance to find the distance between two videos. If they are close, it is regarded that the demonstrations fall into the same task. Therefore, the demonstrations from the same task are viewed as autonomously-collected data that can be used to be trained in different tasks.

In robotic manipulation tasks, synthetic data as input and separate networks in training are RL-based ideas to equip robotic arms with transferability. Moreover, MAML with imitation learning methods do well in task and domain transfer with relatively few data. In the future, training with unlabeled data will be a trend, at that time, autonomous systems need to label the training data by means of unsupervised methods. On the other hand, the testing demonstrations in meta-imitation learning will be extremely fewer, even with no demonstrations, so that an accurate and transferable model is needed.

\section{Discussion and Future works} \label{4}
This review shows the powerful effects of traditional DL, adversarial learning, RL and meta-learning on complex visual and control tasks in autonomous systems. In particular, some traditional DL-based methods may not guarantee the accuracy when transferred to another domain, however, adversarial learning, RL and meta-learning are able to treat transferability well. Although adversarial learning, like GANs, produce better, clearer, and more transferable results than other traditional DL-based methods; Meta-learning methods or combining them with RL and imitation learning methods tend to be equipped with an efficiency or transferability or both of them.

\subsection{Discussion}

In this review, we introduce several typical perception and decision-making tasks of autonomous systems from the perspectives of accuracy and transferability. Since autonomous systems may have a better perception accuracy under good lighting environments than some harsh environments, we first introduce image style transfer, which can change the training data from night to day, rain to sunny, etc. Moreover, image style transfer can realize the transfer of synthetic datasets to real-world images, which greatly reduces the hardware loss caused by directly using autonomous systems for real scenes. Then we review image enhancement and image restoration. Autonomous systems usually involve tasks like image SR and image deblurring/dehazing/rain removal. We review recent developments from the perspectives of accuracy and transferability. When the image quality reaches a good perceptible state, we consider two typical high-level perception tasks of autonomous systems, i.e., semantic segmentation and depth estimation. Since these two tasks are difficult to obtain the ground truth labels, various methods have been proposed for the transferability between the synthetic datasets and real-world data. In addition, we review the tasks of pedestrian detection, person re-ID and pedestrian tracking that are often involved in autonomous systems. Among them, pedestrian detection mainly aims to improve the accuracy of the detection results; person re-ID is a similar task to pedestrian detection but more difficult, which requires matching pedestrians in disjoint camera views; pedestrian tracking pays attention to transferring network and video features and promoting frameworks accuracy. Furthermore, we consider two perception and decision-making tasks of robotic systems, i.e., robot navigation and robotic manipulation. Robot navigation tasks focus on accurately learning task features and transferring information across tasks or domains by RL or meta-learning. Robotic manipulation deals with domain-transfer tasks with robotic arm control more precisely. These two tasks have simulation platforms or practice platforms or both to verify the accuracy or transferability.

\subsection{Future works}

There are still important challenges and future works worth our attention. In this subsection, we summarize some trends and challenges for autonomous systems.
\begin{itemize}
	\item \textbf{GANs with good stability, quick convergence and controllable mode.}  GANs employ the gradient descent method to iterate the generator and discriminator to solve the minimax game problem. In the game, the mutual game between the generator and discriminator may cause that model training unstable, difficult to converge, and even mode collapses. Although there are some preliminary studies aiming at improving these deficiencies of GANs \cite{che2016mode}, \cite{ghosh2018multi}, there is still much room for improvement in terms of the modal diversity and real-time performance. In addition, controlling the mode of data enhancement is still an open question. How to make the generated data mode controllable by controlling additional conditions and keep the model stable, and to achieve purposeful data enhancement, in particular for the computer vision tasks in autonomous systems, is an interesting direction in the future.
	\item \textbf{GANs for complex multi-task.} Although GANs have achieved great results in some typical computer vision tasks of autonomous systems, it still remains difficult to consider the development of more complex multi-task in the future. Since some visual tasks are often related to each other, this phenomenon makes it possible to seamlessly reuse supervision between related tasks or solve different tasks in one system without adding complexity \cite{zamir2018taskonomy}. For example, it is promising to consider training a general-purpose network that can be used for multi-task image restoration in a bad weather condition with only fine-tuning, such as image rain removal, snow removal, dehazing, seasonal change, light adjustment, etc. In addition, in severe rain and fog weather, how to perform image SR while removing rain/dehazing at the same time is challenging. In short, the use of GANs for more complex multi-task remains an open question, and worth exploring.
	\item \textbf{GANs for more challenging domain adaptation.} In autonomous systems, transferability is important for computer vision tasks. Although some results introduce GANs into domain adaptation to improve domain transfer \cite{atapour2018real}, \cite{bousmalis2017unsupervised}, there is still much room for development. When considering more diverse domains, more differentiated cross-domain, and cross-style domains, such as road scenarios in different countries, the existing methods often cannot guarantee good transferability among these domains. However, GANs are promising to develop more diverse domain adaptations by showing unprecedented effectiveness in domain transfer. It is interesting to study the further use of GANs for more differentiated cross-domain transferability.
    \item \textbf{RL for multi-modal, multi-task and multi-agent.}
    Most of RL methods in applications focus primarily on visual input only. However, when considering information from multiple models, such as voice, text, and video, agents can better understand the scenes and the performance in experiments will be more accurate and satisfactory \cite{anderson2018vision}, \cite{wang2019reinforced}. Moreover, in multi-task RL models, the agent is simultaneously trained on both auxiliary tasks and target tasks \cite{bellemare2019geometric}, \cite{jaderberg2016reinforcement}, so that the agent has the ability to transfer experience between tasks. Furthermore, thanks to the distributed nature of the multi-agent, multi-agent RL can achieve learning efficiency from sharing experience, such as communication, teaching and imitation, etc \cite{bucsoniu2010multi}.

    \item \textbf{Meta-learning for unsupervised tasks.}
    Traditional meta-learning, that is, supervised learning during training and testing, in which both training data and testing data are labelled. However, if we use the unlabeled training data, in other words, there is no reward generated in training, how can we also achieve better results on specific tasks during testing? Leveraging unsupervised embeddings to automatically construct tasks or losses for unsupervised meta-learning is a solution \cite{gupta2018unsupervised}, \cite{wortsman2019learning}, \cite{hsu2018unsupervised}. After that, the training tasks for meta-learning are constructed. Therefore, meta-learning issues can be transformed into a wider unsupervised application. It is interesting to use unsupervised meta-learning methods in more realistic task distributions so that the agent can explore and adapt to new tasks more intelligently, and the model can solve real-world tasks more effectively.

    \item \textbf{The application performance of RL and meta-learning.}
    In order to deal with the differences between simulation environments and real scenes, the tasks or the networks can be transferred successfully using RL or meta-learning. Chances are that most of the existing algorithms with good performance in simulation cannot perform as well in real world \cite{li2017deep}, which limits the applications of the models in simulation. Therefore, content-rich and flexible simulation frameworks, like physics engines such as AI2-THOR \cite{zhu2017target}, MuJoCo \cite{todorov2012mujoco}, GymFC \cite{koch2019reinforcement} or like synthetic datasets, such as SUNCG \cite{zhang2017deep} and like robot operating platforms, such as V-REP (virtual robot experiment platform) \cite{rohmer2013v} will help to keep the learned information in more details so that when transferred in real world, the performance is possibly good  \cite{finn2017one}, \cite{yu2018one}. 
    In the future, more informative simulation environments and more flexible real platforms will shorten the gap between simulation and real world, thereby making the model more accurate and transferable. For example, humanoid robotic hand with multiple degrees of freedom is able to deal with tasks more accurately \cite{andrychowicz2020learning}; 3D simulation involving shared tasks, simulators and datasets, assures that the learned skills can be transferred successfully to reality \cite{savva2019habitat}, etc.
    %, is able to be trained entirely in simulation and to be transferred successfully in real world. 
    In a word, due to the high similarity between simulation and real-world platforms, various high-complexity applications trained in simulation can be accurately transferred into practice.

%	\item \textbf{Meta-learning has severe requirements for training.}
%The training of the meta-learning models require thousands of trials and errors to train iteratively. In order to equip the model with the ability to learn new tasks or in new domain fast, the training tasks is much more various and heavy in calculation compared to RL. Therefore, the server need to be updated and expanded to withstand so much data.

\end{itemize}

\section{Conclusion} \label{5}

In this review, we aim to contribute to the evolution of autonomous systems by exploring the impacts of accuracy or transferability or both of them on complex computer vision tasks and decision-making problems. To this end, we mainly focus on basic challenging perception and decision-making tasks in autonomous systems, such as image style transfer, image SR, image deblurring/dehazing/rain removal, semantic segmentation, depth estimation, pedestrian detection, person re-ID, pedestrian tracking, robot navigation and robotic manipulation, etc. We introduce some basic concepts and methods of transfer learning and its special case domain adaptation. Then, we briefly discuss three typical adversarial learning networks, including GANs, cGANs, and CycleGAN. We also present some basic concepts of RL, explain the idea of meta-learning, and discuss the relationship between adversarial learning, RL and meta-learning. Additionally, we analyze some typical DL methods and focus on the powerful performance of GANs in computer vision tasks, discuss RL and meta-learning methods in robot control tasks in both simulation and real-world. 
Moreover, we provide summary tables of learning-based methods for different tasks in autonomous systems, which include the training manners, loss function of models and experiment platforms in visual and robot control tasks. Finally, we discuss main challenges and future works from the aspects of perception and decision-making of autonomous systems by considering the accuracy and transferability.

%\section{Acknowledgements}
%The authors would like to express their sincere appreciation to the
%Associate Editor and the anonymous reviewers for their insightful
%comments.

%\section{Acknowledgements}

\section{Author contributions}
Ideas design, Y. Tang and C. Zhang; Reference Collection, Y. Tang, C. Zhang and J. Wang; Writing-Original Draft, C. Zhang and J. Wang; Writing-Review, Y. Tang, G.Yen, C. Zhao, Q. Sun and J. Kurths.

  \label{}



%\nocite{*}
\bibliographystyle{IEEEtran}
\bibliography{Ref}% Produces the bibliography via BibTeX.

%\begin{IEEEbiography}
%[{\includegraphics[width=1in,height=1.5in,clip,keepaspectratio]{figures/Tang.eps}}]{Yang
%Tang}

%(M'11) received the B.S. degree and Ph.D. degree in electrical
%engineering from Donghua University, Shanghai, China in 2006 and
%2010, respectively. He was a Research Associate with The Hong Kong
%Polytechnic University, Kowloon, Hong Kong, China, from 2008 to
%2010. He was an Alexander von Humboldt Research Fellow with the
%Humboldt University of Berlin, Berlin, Germany from 2011 to 2013. He
%was a Visiting Research Fellow with Brunel University, London, U.K.
%in 2012. He has been a Research Scientist with the Potsdam Institute
%for Climate Impact Research, Potsdam, Germany and a Visiting
%Research Scientist with the Humboldt University of Berlin, Berlin,
%Germany since 2013. He has published more than 30 refereed papers in
%international journals. His main research interests are
%synchronization/consensus, networked control systems, evolutionary
%computation, bioinformatics and their applications. Dr. Tang is an
%Associate Editor for Neurocomputing and a Guest Editor for Journal
%of the Franklin Institute.

%\end{IEEEbiography}

\end{document}